\documentclass[lettersize,journal]{IEEEtran}
\usepackage{booktabs}
\usepackage{array}
\usepackage[caption=false,font=normalsize,labelfont=sf,textfont=sf]{subfig}
\usepackage{cite}
\usepackage{amsmath,amssymb,amsfonts}
\usepackage{graphicx}
\usepackage{textcomp}
\usepackage{xcolor}
\usepackage{url}
\usepackage{verbatim}
\usepackage{multirow}
\usepackage{bigstrut}
\usepackage[figuresright]{rotating}
\usepackage{makecell}
\usepackage{float}
\usepackage{colortbl}
\usepackage{utfsym}
\usepackage{bbding}
\usepackage{pifont}
\usepackage{algorithm}
\usepackage{algorithmic}
\usepackage{relsize}
\usepackage[dvipsnames]{xcolor}
\usepackage{nicefrac}
\usepackage[colorlinks=true, citecolor=blue, linkcolor=blue, urlcolor=blue]{hyperref}
\begin{document}

\title{Attentive Convolution: Unifying the Expressivity of Self-Attention with Convolutional Efficiency}

\author{
Hao Yu, Haoyu Chen, Yan Jiang, Wei Peng, Zhaodong Sun, Samuel Kaski, Guoying Zhao, \IEEEmembership{Fellow,~IEEE}

\thanks{This work was supported by the Research Council of Finland Academy Professor project EmotionAI (grants 336116, 345122, 359854), the University of Oulu \& Research Council of Finland Profi 7 (grant 352788), and Academy of Finland Flagship program: the Finnish Center for Artificial Intelligence FCAI. (Corresponding Author: Guoying Zhao)}

\thanks{H. Yu, H. Chen, Y. Jiang, G. Zhao are with Center for Machine Vision and Signal Analysis, University of Oulu, Finland (e-mail: hao.2.yu@oulu.fi; chen.haoyu@oulu.fi; yan.jiang@oulu.fi; guoying.zhao@oulu.fi).}

\thanks{W. Peng is with the Department of Psychiatry and Behavioral Sciences, Stanford University, USA. (e-mail: wepeng@stanford.edu).}

\thanks{Z. Sun is with the School of Computer Science, Nanjing University of Information Science and Technology, China (e-mail: zhaodong.sun@nuist.edu.cn).}

\thanks{S. Kaski is with the Department of Computer Science, Aalto University, Finland. (e-mail: samuel.kaski@aalto.fi).}

\thanks{S. Kaski and G. Zhao are with the ELLIS Institute Finland (first affiliation).}

}

\markboth{Journal of \LaTeX\ Class Files,~Vol.~14, No.~8, August~2021}%
{Shell \MakeLowercase{\textit{et al.}}: A Sample Article Using IEEEtran.cls for IEEE Journals}


\maketitle

\begin{abstract}
Self-attention (SA) has become the cornerstone of modern vision backbones for its powerful expressivity over traditional Convolutions (Conv). However, its quadratic complexity remains a critical bottleneck for practical applications. Given that Conv offers linear complexity and strong visual priors, continuing efforts have been made to promote the renaissance of Conv. However, a persistent performance chasm remains, highlighting that these modernizations have not yet captured the intrinsic expressivity that defines SA. In this paper, we re-examine the design of the CNNs, directed by a key question: what principles give SA its edge over Conv? As a result, we reveal two fundamental insights that challenge the long-standing design intuitions in prior research (e.g., Receptive field). The two findings are: (1) \textit{Adaptive routing}: SA dynamically regulates positional information flow according to semantic content, whereas Conv employs static kernels uniformly across all positions. (2) \textit{Lateral inhibition}: SA induces score competition among token weighting, effectively suppressing redundancy and sharpening representations, whereas Conv filters lack such inhibitory dynamics and exhibit considerable redundancy. Based on this, we propose \textit{Attentive Convolution} (ATConv), a principled reformulation of the convolutional operator that intrinsically injects these principles. Interestingly, with only $3\times3$ kernels, ATConv consistently outperforms various SA mechanisms in fundamental vision tasks. Building on ATConv, we introduce AttNet, a CNN family that can attain \textbf{84.4\%} ImageNet-1K Top-1 accuracy with only 27M parameters. In diffusion-based image generation, replacing all SA with the proposed $3\times 3$ ATConv in SiT-XL/2 reduces ImageNet FID by 0.15 in 400k steps with faster sampling. Code is available at: github.com/price112/Attentive-Convolution.
\end{abstract}


\section{Introduction}

Convolutional neural networks (CNNs)~\cite{he2016deep, simonyan2014very, krizhevsky2012imagenet, lecun1998convolutional, huang2017densely, szegedy2015going} have long dominated computer vision, achieving remarkable success across diverse tasks owing to their inherent visual inductive biases and computational efficiency on high-dimensional inputs. Recently, Vision Transformers (ViTs) \cite{dosovitskiy2020image, touvron2021training, liu2021swin, ali2021xcit, dong2022cswin, wang2021pyramid} have emerged as a strong competitive alternative, leveraging the self-attention mechanism \cite{vaswani2017attention} to model global dependencies through content-based query–key interactions. Unlike convolutions, which encode fixed local patterns with limited receptive fields, self-attention enables flexible long-range modeling. However, vanilla self-attention suffers from quadratic computational complexity with respect to the input resolution, rendering it inefficient for visual data characterized by high dimensionality and substantial encoding redundancy. Despite that the ViT series \cite{dosovitskiy2020image} leverages large patch sizes (e.g., 16x16 or 32x32) to reduce the sequence length of input images, its computational complexity is still much higher than that of traditional CNNs. Furthermore, self-attention's position-agnostic design treats all spatial locations equally during pixel dependency modeling, requiring extensive training resources to learn fundamental visual priors such as locality and object continuity from scratch.

To address these limitations, modern ViTs have undergone substantial architectural evolution. An important revolution was the adoption of CNN-like pyramid designs \cite{wang2021pyramid, liu2021swin}, which progressively downsample spatial dimensions, restricting attention computation to manageable resolutions. Based on this, recent state-of-the-art ViTs \cite{vasu2023fastvit, shaker2023swiftformer, wang2022pvt, shi2024transnext,wu2022p2t} further hybridize attention with depth-wise convolutions \cite{chollet2017xception, howard2017mobilenets}, creating models that strategically leverage convolution's efficiency for local processing while preserving attention's capacity for global modeling. Notably, the persistence of convolution reveals a crucial insight: its visual efficiency and inherent visual inductive biases remain indispensable for vision systems.

Recognizing this complementarity, researchers have sought to augment CNNs with ViT-inspired principles. The most common idea is to expand the receptive field that approximates the global modeling of attention. ConvNeXt \cite{liu2022convnet, woo2023convnext} modernizes ConvNets with ViT design elements and adopts larger kernels instead of the traditional $3\times3$ ones, while reparameterization-based architectures \cite{ding2021repvgg, ding2022scaling} further extend kernel sizes up to 31$\times$31 to capture long-range dependencies. Although these strategies yield notable improvements over classical CNNs such as ResNet \cite{he2016deep}, a substantial gap from modern ViTs persists. For example, with similar parameters, TransNeXt-Tiny \cite{shi2024transnext} wins ConvNeXt-Tiny \cite{liu2022convnet} by 2.0\% Top-1 accuracy in ImageNet-1K classification. This disparity highlights that simply augmenting CNN from the receptive field perspective is insufficient to capture the fundamental advantages of ViTs.

\begin{figure*}[t]
  \centering
   \includegraphics[width=1.0\linewidth]{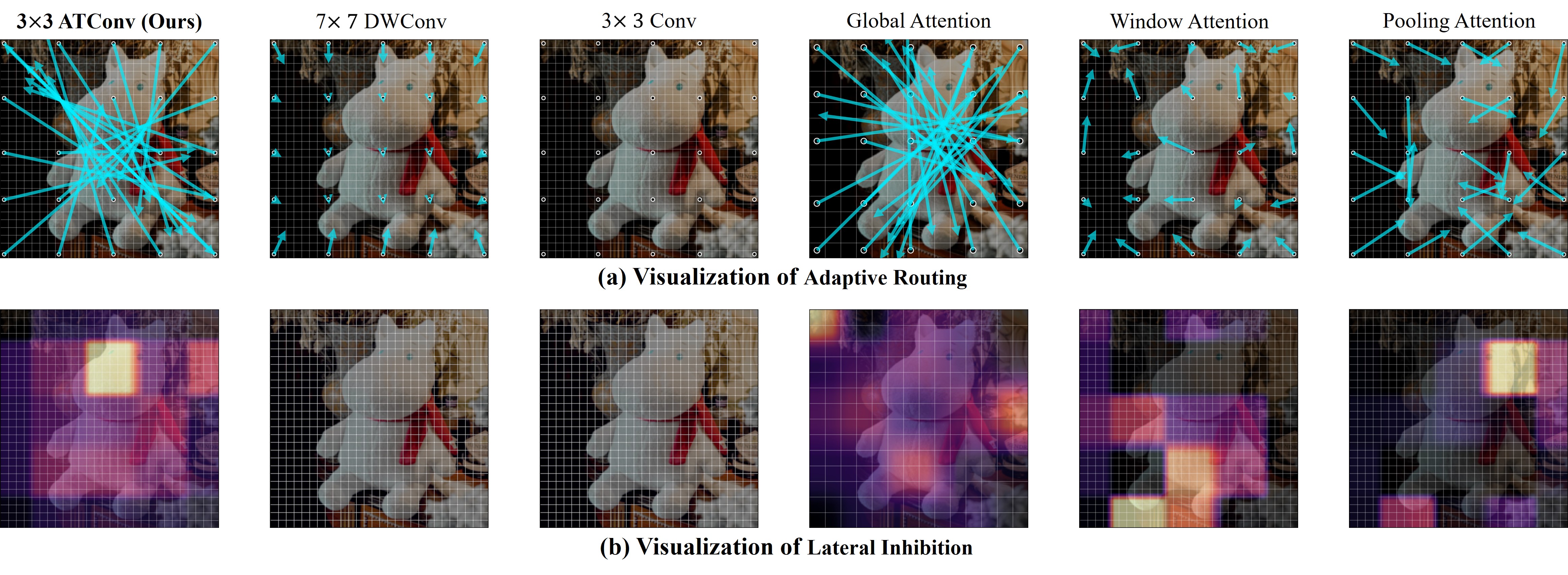}
   \captionsetup{skip=2pt}
\caption{ Visualization of adaptive routing and lateral inhibition on 3$\times$3 ATConv, 7$\times$7 DWConv \cite{liu2022convnet}, 3$\times$3 Conv \cite{he2016deep}, vanilla global attention \cite{dosovitskiy2020image}, window \cite{liu2021swin} and pooling \cite{wang2021pyramid} attention. We analyze operator-intrinsic behavior by extracting influence maps $G(h,w)=\sum_c \lvert \partial y_{c^\ast,h^\ast,w^\ast}/\partial x_{c,h,w}\rvert$. Adaptive routing is visualized via distance-weighted centroids of high-influence regions, with $\mathrm{FAR}$ at radial threshold $r_0$ measuring long-range preference. Diverse arrows indicate different inputs produce different aggregation patterns, while uniform arrows reveal fixed routing. Lateral inhibition is quantified by the off-center suppression response to positive perturbations at anchors, where brighter regions indicate stronger surround suppression.  \textit{ATConv with a compact 3$\times$3 kernel exhibits pronounced adaptive routing and lateral inhibition effects similar to the global self-attention}. See our depository for detailed visualization code.}
\label{fig:1}
\vspace{-1.5em}
\end{figure*}

\IEEEpubidadjcol

In this paper, we dive deeper and present two new perspectives to understand the underlying visual expressivity of self-attention over Conv-based operators: the \textit{adaptive routing} and \textit{lateral inhibition} properties. Through a unified weighted aggregation framework, we demonstrate that vanilla convolution is a static weight aggregation operator with an identity representation basis, which leverages the fixed pixel aggregation rule across all spatial positions. In contrast, self-attention implements \textit{adaptive routing} via query-key interactions with learnable basis space transformations. This adaptivity enables attention modeling to have selective information transfer based on semantic relevance rather than fixed rules. More importantly, the softmax normalization over attention scores induces crucial \textit{lateral inhibition} dynamics \cite{blakemore1970lateral} within attention calculation, a canonical mechanism of biological vision where neurons compete for visual selection. In the primary visual cortex (V1) \cite{tootell1998functional}, neurons exhibit similar center-surround antagonism, with neighboring neurons competing to represent different features. This competitive dynamic prevents redundant encoding and sharpens feature selectivity—precisely the characteristics lacking in standard convolution. In self-attention mechanisms, the softmax on attention score creates competitive dynamics which implements mutual suppression: increasing the attention weight to one position necessarily decreases others. This sharpens the representations and avoids giving flat responses (e.g., low-rank prediction). The resulting competitive pressure amplifies discriminative connections while suppressing spurious or noisy patterns, leading to more selective and robust representations. In contrast, convolution’s static and independent kernels lack such inhibitory dynamics, fundamentally constraining their expressivity and leading to substantial representation redundancy.

To validate our analysis, we propose \textit{Attentive Convolution} (ATConv), a principled revolution of convolution that embeds the adaptive routing and lateral inhibition principles distilled from self-attention. Unlike vanilla convolutions that employ static, input-agnostic kernels, ATConv adaptively derives its kernels from the input through a context-to-kernel translation mechanism. The key insight is that although convolution operates locally, its kernels can be contextually conditioned to encode global dependencies. In this way, global semantic relevance is translated into input-dependent kernels, implicitly establishing adaptive routing correspondences through local operations, without incurring the quadratic cost of global pairwise calculation. Beyond routing, we further inject the differential kernel modulation (DKM) into the operation logic of ATConv, which is designed to introduce the lateral inhibition dynamic tailored for the convolutional framework. Inspired by classical descriptors such as LBP \cite{ojala2002multiresolution, zhao2007dynamic} and DoG \cite{marr1980theory}, the DKM mechanism performs difference-oriented kernel modulation during convolutional operation, dynamically enhancing feature sharpness while suppressing redundant channel responses.

Fig.~\ref{fig:1} empirically validates our findings. In part (a), routing arrows depict how each position aggregates information: uniform arrows denote fixed routing, while diverse arrows indicate adaptive routing for different pixel positions. Traditional Conv operators (3$\times$3 Conv and 7$\times$7 DWConv) show a fixed local routing. Even with a larger kernel, 7$\times$7 DWConv yields identical aggregation patterns across positions, underscoring that kernel enlargement cannot induce adaptivity. In contrast, global self-attention produces diverse routing across spatial regions, while pooling and window attention trade routing capacity for efficiency due to local attention constraints. In contrast, ATConv achieves adaptive routing comparable to global self-attention using only compact 3$\times$3 kernels. Its diverse arrow patterns confirm that the context-to-kernel translation encodes global correspondences into local kernels, enabling convolutional traversal to realize global adaptive routing with even a compact 3$\times$3 kernel size.

Part (b) visualizes the lateral inhibition effect. Standard Conv and DWConv show no inhibitory behavior, highlighting the absence of neuron suppression in traditional convolutional operators. In contrast, ATConv demonstrates competitive dynamics akin to the three compared attention mechanisms, with strong gradient-score competition that sharpens representations and suppresses redundancy.

Building on ATConv, we introduce the Attentive Convolutional Network (AttNet), a purely convolutional architecture that attains state-of-the-art performance while dispensing with self-attention. Moreover, we show that ATConv can serve as a drop-in replacement for self-attention, consistently improving accuracy and efficiency across diverse vision tasks and backbones. Our main contributions are as follows.

- We identify \textit{adaptive routing} and \textit{lateral inhibition} as the key mechanisms behind the superior expressivity of self-attention, and provide both theoretical and empirical evidence that these properties govern representational expressivity.

- We propose \textit{Attentive Convolution} (ATConv), which embeds the adaptive routing and lateral inhibition principles into the convolutional framework, delivering attention-level expressivity with convolutional efficiency.

-We evaluate ATConv across a wide range of discrimination and generation tasks, revealing its consistent advantages over leading attention mechanisms. These results solidify ATConv's position as a new foundational operator, poised to drive the development of next-generation visual models.

\section{Related Work}
\subsection{Vision Transformer}
In the field of natural language processing, Transformer \cite{vaswani2017attention} leverages self-attention to model global dependencies between tokens. Vision Transformer (ViT) \cite{dosovitskiy2020image} pioneered the adaptation of this mechanism to computer vision, demonstrating that pure attention-based architectures can achieve competitive performance on image recognition tasks. However, the quadratic complexity of self-attention and the lack of visual inductive biases bring significant computational overhead in visual tasks. The adaptation of self-attention in visual data has motivated extensive research along the following two primary directions.

\textbf{Computational efficiency.} The $O(N^2)$ complexity of self-attention becomes prohibitive for high-resolution inputs like natural images. Pooling-based methods (e.g., PVT \cite{wang2021pyramid}, PVTv2 \cite{wang2022pvt}, P2T \cite{wu2022p2t}) reduce computation through spatial downsampling. Window-based methods restrict attention scope, like Swin Transformer \cite{liu2021swin} employs shifted windows for linear complexity. Afterwards, more advanced window partition techniques brought further performance improvements, e.g., Cswin \cite{dong2022cswin} uses cross-shaped windows for efficient global modeling, and MaxViT \cite{tu2022maxvit} leverages the block-grid interlaced window to promote information flow of local attention. Linear attention variants fundamentally alter the computational paradigm by replacing softmax with kernel functions, exploiting associative properties to achieve $O(N)$ complexity. XCiT \cite{ali2021xcit} computes cross-covariance between feature channels rather than token-wise attention. InLine attention \cite{han2024bridging} introduces injectivity constraints to preserve discriminative power without using softmax, and CosFormer \cite{qin2022cosformer} incorporates cosine-based reweighting for improved stability.

\textbf{Inductive biases.} Self-attention lacks the visual priors inherent to convolutions, necessitating explicit incorporation via positional encodings or Conv integration. Positional encodings provide crucial spatial awareness through various formulations (relative \cite{shaw2018self}, 2D RoPE \cite{heo2024rotary}, CPB \cite{liu2022swin}, LePE \cite{graham2021levit}). More effective approaches directly combine convolutions: CoAtNet \cite{dai2021coatnet} systematically integrates depth-wise convolution with self-attention, while CvT \cite{wu2021cvt} introduces convolutional token embedding and projection layers. InLine \cite{han2024bridging} attention systematically explains the importance of convolutional locality in attention modeling. To further improve efficiency, FastViT \cite{vasu2023fastvit} builds a more progressive CNN-ViT hybrid architecture only by using attention in the last stage. Biologically-inspired architectures leverage human vision principles, e.g., Focal Transformer implements coarse-to-fine attention across resolutions, while TransNeXt \cite{shi2024transnext} aggregates multiscale local features through pixel-focused attention.

While extensive efforts have focused on augmenting self-attention with convolutional biases, the dual direction, enhancing convolutions with attention's key advantages, remains underexplored. Since convolution inherently provides visual efficiency, augmenting it with attention's positive principles present a promising direction. This paper shows that transferring self-attention's positive principles to Conv can significantly improve performance.

\subsection{Convolutional Neural Networks}
The convolutional operator \cite{lecun1989handwritten} lies at the heart of modern visual recognition models. By sliding a shared kernel across local neighborhoods, it imposes several strong \textit{visual inductive biases}, including locality, translation equivariance, weight sharing, and hierarchical feature composition. These properties align well with the statistics of natural images. Built on the convolutional operator, Convolutional Neural Networks (CNNs) \cite{he2016deep,howard2017mobilenets,huang2017densely,ding2021repvgg,simonyan2014very,krizhevsky2012imagenet,szegedy2015going} have therefore dominated the field of computer vision for decades, evolving from early models such as LeNet \cite{lecun1989handwritten} to modern architectures such as ResNet \cite{he2016deep}, DenseNet \cite{huang2017densely}, and EfficientNet \cite{tan2019efficientnet}.  
However, with the advent of self-attention in vision tasks, CNNs have been rapidly supplanted by Vision Transformers (ViTs), which exhibit superior performance through the incorporation of self-attention mechanisms with long-range adaptive routing and competitive score modeling capacities.

Representative efforts to narrow down the gap between CNNs and ViTs include ConvNeXt \cite{liu2022convnet}, which adopts architectural paradigms from ViTs and enlarges convolutional kernels from $3 \times 3$ to $7 \times 7$ in order to emulate the long-range modeling capacity of self-attention. Although ConvNeXt shows promising improvements, a substantial performance gap remains compared to the leading ViTs. Subsequent CNNs have similarly emphasized architectural modifications to mimic ViT characteristics, such as employing large kernels (e.g., RepLKNet \cite{ding2022scaling}, InceptionNeXt \cite{yu2024inceptionnext}), dilated convolutions \cite{yu2025freenet}. However, these efforts primarily address receptive-field limitations while overlooking a more fundamental issue: the gap stems not merely from spatial coverage, but from the intrinsic differences in visual modeling between convolution and self-attention. This paper steps further by analyzing the key attributes of self-attention in visual modeling and reforging the Conv operator accordingly, aiming to further mitigate the performance gap between CNNs and ViTs.

\subsection{Dynamic Convolutional Architectures}
Early explorations into content-adaptive convolutions aimed to make kernels input-dependent over static schemes. CondConv \cite{yang2019condconv} and DynamicConv \cite{chen2020dynamic} learn mixtures of base kernels, while WeightNet \cite{ma2020weightnet } and ODConv \cite{li2022omni} use hypernetworks to generate dynamic weights. Involution \cite{li2021involution} designs spatially-specific yet channel-agnostic filters. SENet \cite{hu2018squeeze} and CBAM \cite{woo2018cbam} recalibrate features with attention modules. While these approaches outperform vanilla convolutions, they typically incur high computational/parameter costs and lack the performance scalability needed to serve as fundamental backbone operators. Consequently, most dynamic convolutional designs remain as auxiliary modules rather than being core components of modern architectures.

A deeper limitation of these methods is the absence of \emph{lateral inhibition} \cite{blakemore1970lateral}, the competitive dynamic central to self-attention. Existing dynamic convolutions typically modulate features independently via gates or additive combinations, without enforcing inter-feature competition. This leads to diffuse responses unable to suppress noise or irrelevant signals, explaining why even complex dynamic operators still lag far behind self-attention. ATConv addresses these gaps by incorporating \emph{adaptive routing} to capture content-dependent adaptation and \emph{lateral inhibition} to introduce competitive dynamics. This synergy produces sharper, more discriminative representations reminiscent of self-attention, yet preserves the efficiency and structural simplicity of convolution. Importantly, ATConv's verified generality, efficiency, and scalability establish it as a strong candidate for a foundational operator similar to the self-attention, transcending the limitations of prior dynamic-based approaches.

\section{Methodology}
\subsection{Preliminaries}
To analyze the intrinsic difference between self-attention and Conv, we first establish a framework where both operators perform weighted aggregation over a signal manifold. Let $\mathbf{X}\in \mathbb{R}^{B\times C\times H\times W}$ denote the input signal with spatial dimensions $H \times W$ and feature (channel) dimension $C$. For analysis, we also use the flattened view $\mathbf{X}\in \mathbb{R}^{B\times N\times C}$ with $N=H\times W$.

\noindent\textbf{Definition 1: Generalized Aggregation Operator.} Given the input signal $\mathbf{X}$, the output at position $i$ is:
\begin{equation}
    \mathbf{y}_i = \sum_{j\in\Omega_i} \alpha_{ij} \cdot \mathcal{T}(\mathbf{X}_j),
    \label{eq:1}
\end{equation}
where $\Omega_i$ denotes the aggregation domain, $\alpha_{ij}$ represents the aggregation weight from position $j$ to position $i$, and $\mathcal{T}: \mathbb{R}^{C}\rightarrow \mathbb{R}^{C'}$ is the basis transform at position $j$. The crucial distinction between self-attention and Conv lies in how $\alpha_{ij}$ and $\mathcal{T}_j$ are determined. In the following, we analyze Conv in its \emph{depthwise} form, which uses a scalar weight per spatial offset for each channel and is more comparable to self-attention’s content-weighted summation.

\begin{figure*}[t]
  \centering
   \includegraphics[width=0.95\linewidth]{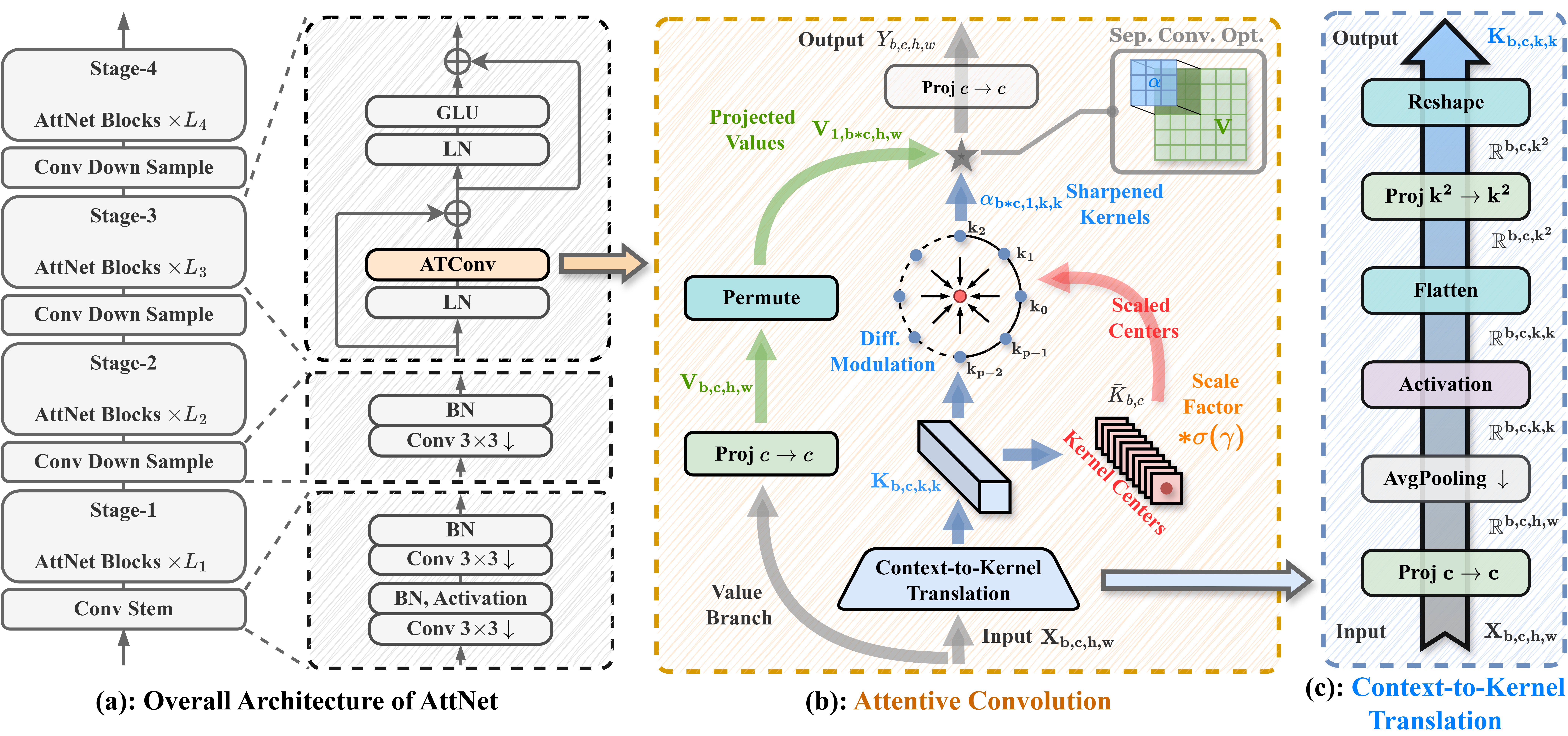}
   \captionsetup{skip=0pt}
   \caption{(a) Overall architecture of AttNet, where we use ATConv as token-mixer (spatial-operator) and GLU as channel-mixer; (b) Architecture of Attentive Convolution (ATConv); (c) Architecture of kernel generator which builds the initial kernels for ATConv based on input contents.}
   \label{fig:2}
\end{figure*}

\subsection{The Fundamental Distinction in Signal Aggregation}
\label{sec:2b}
\noindent\textbf{Adaptive Routing vs. Static Aggregation.} 
We identify adaptive routing as the fundamental distinction between self-attention and convolution. At its core, \textit{routing} refers to how information flows from input positions to output during pixel aggregation, encompassing both which positions contribute ($\alpha_{ij}$) and what representations they provide ($\mathcal{T}_j$). Our insight is that achieving adaptive routing requires both components to adapt dynamically based on input content.

Conv employs \textit{static routing}, where aggregation are fixed regardless of input:
\begin{equation}
    \mathbf{y}_i^{\text{Conv}} = \sum_{j\in\mathcal{N}_K(i)} w_{p(i,j)}\;\mathbf{x}_j,
    \label{eq:2}
\end{equation}
here, $w_{p(i,j)}$ denotes the weight of the kernel in relative position $p(i,j) = j-i$, and $\mathcal{N}_K(i)$ represents the local neighborhood $K\times K$. These weights remain input-independent ($\nicefrac{\partial w_{p(i,j)}}{\partial \mathbf{X}}=\mathbf{0}$), and the basis features undergo no transformation mapping before aggregation ($\mathcal{T}_j^{\text{Conv}}(\mathbf{X}) = \mathbf{x}_j$). Consequently, convolution applies identical aggregation patterns universally: a 3$\times$3 edge detector employs the same weights whether processing edges, textures, or uniform regions, fundamentally limiting its adaptability.

Self-attention, instead, achieves \textit{adaptive routing} through two synergistic mechanisms:
\begin{equation}
\begin{aligned}
    \mathbf{y}_i^{\text{SA}} &= \sum_{j=1}^{N} \alpha_{ij}^{\text{SA}}(\mathbf{X}) \cdot \mathbf{v}_j, \\
    \alpha_{ij}^{\text{SA}} &= \frac{\exp(s_{ij}/\tau)}{\sum_{m=1}^{N} \exp(s_{im}/\tau)},
    \label{eq:3}
\end{aligned}
\end{equation}
where $s_{ij} = \langle \mathbf{q}_i, \mathbf{k}_j \rangle$ quantifies query-key affinity. The routing weights $\alpha_{ij}^{\text{SA}}$ adapt dynamically based on content similarity, while the value $\mathbf{v}_j = \mathbf{W}_v\mathbf{x}_j$  (the $\mathcal{T}$ in Eq.~\ref{eq:1}) projects basis features into an optimized representation space. This dual adaptation makes weights determine \textit{where} to aggregate based on semantic relevance, while value projection determines \textit{what} representations to aggregate for better information routing.

To further establish why both components are necessary, we analyze the sensitivity of each operator to input perturbations with the Kronecker delta $\delta_{jn}$:
\begin{equation}
\begin{aligned}
    \frac{\partial \mathbf{y}_i^{\text{Conv}}}{\partial\mathbf{x}_n} &= \begin{cases}
        w_{p(i,n)} \cdot \mathbf{I}_{C'} & \text{if } n \in \mathcal{N}_K(i) \\
        \mathbf{0}_{C'} & \text{otherwise},
    \end{cases} \\
    \frac{\partial \mathbf{y}_i^{\text{SA}}}{\partial\mathbf{x}_n} &= \sum_{j=1}^{N} \left[ \frac{\partial \alpha_{ij}^{\text{SA}}}{\partial \mathbf{x}_n} \cdot \mathbf{v}_j + \alpha_{ij}^{\text{SA}} \cdot \delta_{jn} \cdot \mathbf{W}_v \right].
    \label{eq:4}
\end{aligned}
\end{equation}

This sensitivity analysis reveals a fundamental computational hierarchy. The static weights of Conv yield gradients invariant to input content, constraining it to uniform spatial processing on untransformed features. In contrast, self-attention's sensitivity decomposes into two adaptive components that enable content-aware computation. The first term, $(\partial \alpha_{ij}^{\text{SA}} / \partial \mathbf{x}_n)\,\mathbf{v}_j$, facilitates adaptive routing by adjusting aggregation weights based on input characteristics. The second term, $\alpha_{ij}^{\text{SA}} \cdot \mathbf{W}_v$, represents learned value transformations that project features into task-optimized subspaces. This transformation is critical: without it ($\mathbf{W}_v = \mathbf{I}$), the term degenerates to $\alpha_{in}^{\text{SA}} \cdot \mathbf{I}_{C'}$, restricting aggregation to the original feature space where semantic concepts may be poorly separated. Value projections enable discovering task-specific manifolds where similar concepts cluster and dissimilar ones separate, facilitating more effective aggregation.

This analysis establishes a clear progression in routing capabilities. Vanilla Conv implements neither adaptive components, operating with fixed kernels on untransformed features. Existing dynamic Convs advance halfway by generating content-dependent weights, yet still aggregate within the original feature space. Self-attention alone achieves complete adaptive routing through both (i) content-dependent weights that determine where to gather information and (ii) learnable feature space transformations that optimize what representations to aggregate. This progression from static to partial to complete adaptation explains the persistent performance gap between convolution and attention in visual tasks.

\noindent\textbf{Lateral Inhibition: Competitive Dynamics in Aggregation.} Beyond adaptive routing, we analyze the lateral inhibition between aggregation weights. Convolution weights operate independently without mutual influence:
\begin{equation}
    w_{p(i,j)} = \text{constant}, \quad \sum_{j\in\mathcal{N}_K(i)} w_{p(i,j)} \neq 1 \text{ (generally)}.
    \label{eq:5}
\end{equation}

Each kernel weight functions in isolation, modifying one weight does not affect others. This independence means all positions contribute according to fixed weights regardless of their relative importance, potentially aggregating both signal and noise with equal emphasis for different inputs. Furthermore, without competitive dynamics to enforce specialization, independent kernels often converge to similar features, producing redundant filters. Multiple channels may learn nearly identical edge detectors or texture filters, resulting in a low rank transformation where hundreds of parameters encode only dozens of unique patterns. This representational redundancy, exacerbated by the absence of lateral inhibition, yields flat, diffuse responses that waste the network's capacity. 

Self-attention implements competitive dynamics through softmax normalization, creating mutual inhibition between aggregation weights:
\begin{equation}
    \frac{\partial \alpha_{ij}^{\text{SA}}}{\partial s_{ik}} = \begin{cases}
    \alpha_{ij}^{\text{SA}}(1 - \alpha_{ij}^{\text{SA}})/\tau & \text{if } j = k \\
    -\alpha_{ij}^{\text{SA}}\alpha_{ik}^{\text{SA}}/\tau & \text{if } j \neq k\ .
    \end{cases}
    \label{eq:6}
\end{equation}

The negative off-diagonal terms $-\alpha_{ij}^{\text{SA}}\alpha_{ik}^{\text{SA}}$ implement lateral inhibition: increasing the affinity score $s_{ik}$ for one position necessarily decreases weights $\alpha_{ij}$ for all other positions $j \neq k$. This creates a dynamic where positions compete for aggregation bandwidth, forcing each attention score to specialize on distinct patterns rather than redundantly encoding similar features. Unlike convolution's independent weights that often converge to similar filters, this competitive pressure ensures diverse representations across heads and positions. The strongest semantic connections amplify while weaker ones suppress, yielding sharp, non-redundant feature maps with high effective rank. This lateral inhibition mechanism thus enhances both representational quality and robustness, as the network learns complementary features that capture different aspects of the input rather than wasting capacity on duplicate patterns.

\noindent\textbf{Summary.} Adaptive routing and lateral inhibition fundamentally distinguish self-attention from convolution, extending beyond differences in receptive field size. These mechanisms enable self-attention from (1) adaptive information flow that responds to semantic content rather than following fixed spatial patterns, and (2) competitive selection that amplifies task-relevant signals and reduces feature redundancy, while suppressing noise. These properties are hypothesized to contribute to self-attention's strong empirical performance on complex vision tasks requiring selective aggregation and global context integration, motivating the development of convolution operators that incorporate these routing capabilities while preserving computational efficiency.

\subsection{Attentive Convolution: From Theory to Design}

Our analysis shows that the strength of self-attention arises from two principles absent in convolution: \emph{adaptive routing}, which enables content-aware aggregation, and \emph{lateral inhibition}, which enables competitive dynamics. The key challenge is to translate these abstract principles into a convolutional framework. Guided by Eq.~\ref{eq:4} and Eq.~\ref{eq:6}, ATConv instantiates these principles through three principled revolutions based on the vanilla convolution framework: (i) a \emph{context-to-kernel translation} mechanism generating routing weights (kernel) that encode the global semantic understanding into local processing rules; (ii) a learnable \emph{value projection} for basis adaptation; and (iii) a \emph{differential kernel modulation} injecting lateral inhibition between kernel entries via difference-oriented modulation.

\noindent\textbf{Adaptive Routing with Context-to-Kernel Translation.}  
To enable adaptive routing within the convolutional framework, ATConv introduces a \emph{Context-to-Kernel Translation} (C2K) mechanism that technically departs from conventional kernel designs. C2K functions as a semantic compiler that bridges global scene understanding with local processing rules. This is reached by encoding the complete $H\times W$ spatial context into compact semantic representations and translating these into tailored filtering operations. Given an input tensor $\mathbf{X}\in \mathbb{R}^{B\times C\times H\times W}$, the C2K executes the following steps:
\begin{equation}
\label{eq:routing_generation}
\begin{aligned}
\mathbf{F} &= \operatorname{Conv}_{1\times1}(\mathbf{X}), &\mathbf{F} \in \mathbb{R}^{B\times C\times H\times W}&,\\
\mathbf{Z} &= \operatorname{AdaAvgPool}_{K\times K}\!\big(\mathbf{F}\big), &\mathbf{Z} \in \mathbb{R}^{B\times C\times K\times K},\\
\hat{\mathbf{K}} &= \mathbf{W}_{\text{gen}}\cdot\operatorname{Vec}\big(\phi(\mathbf{Z})\big), &\mathbf{\hat{K}} \in \mathbb{R}^{B\times C\times K^2},\\
\mathbf{K} &= \operatorname{Reshape}(\hat{\mathbf{K}}), &\mathbf{K} \in \mathbb{R}^{B\times C\times K\times K}.
\end{aligned}
\end{equation}

The above pipeline proceeds through four key steps. First, a pointwise convolution $\operatorname{Conv}_{1\times1}$ projects the input features into a routing-aware latent space, encoding global contextual information for kernel synthesis. Second, adaptive average pooling $\operatorname{AdaAvgPool}_{K\times K}$ compresses the full $H\times W$ spatial resolution into a compact $K\times K$ representation, where each position corresponds to a kernel coefficient location. Third, after vectorizing ($\mathbb{R}^{K\times K}\rightarrow\mathbb{R}^{K^2}$) this pooled representation via $\operatorname{Vec}(\cdot)$, we apply a channel-shared linear transformation $\mathbf{W}_{\text{gen}}\in \mathbb{R}^{K^2\times K^2}$ with nonlinear activation $\phi$ (e.g., GELU) to translate semantic codes into kernel entities. Finally, the result vector is reshaped to recover the $K\times K$ spatial structure.

This design embodies the core principle of \emph{Context-to-Kernel Translation}: global scene knowledge is first distilled into a compact semantic encoding, then re-expressed as spatially-adaptive filtering operations. When convolution occurs at position $(h,w)$, the generated kernel $\mathbf{K}_{b,c,:,:}$ aggregates local neighborhoods using weights informed by the entire spatial context. As the convolution window traverses the feature map, each location receives a uniquely tailored kernel that captures how its local structure relates to the global scene, effectively establishing an implicit routing network. 

Our empirical results in Fig.~\ref{fig:1} and Tab.~\ref{tab:9}-(b) reveal that C2K enables even compact $3\times3$ kernels to capture long-range dependencies at the $224$px scale. This finding highlights a crucial insight: context-aware kernel generation through C2K provides a more effective and efficient pathway to global modeling, rather than mechanically increasing kernel sizes.

\begin{figure*}[t]
  \centering
\includegraphics[width=0.95\linewidth]{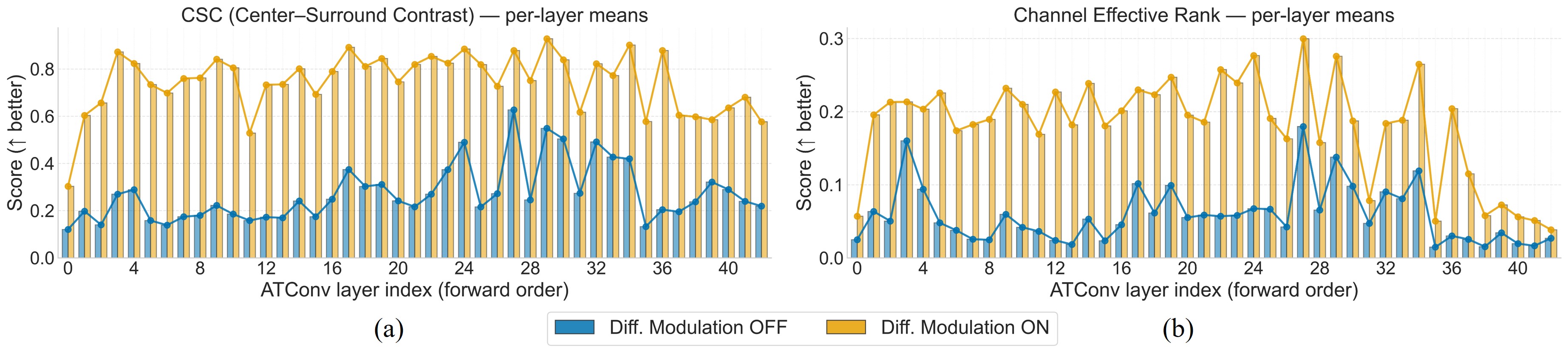}
\captionsetup{skip=1pt}
\caption{Impact of Differential Kernel Modulation (Diff. Modulation) on feature properties. We analyze the output of every ATConv in AttNet-T4 with and without using Diff. Modulation. (a) \textbf{Center-Surround Contrast} is defined as
$\mathrm{CSC}(x) = \mathbb{E}[|x - G_\sigma(x)|]\cdot\mathbb{E}[|x|]^{-1}$,
where $G_\sigma(\cdot)$ is Gaussian blurring. CSC quantifies the relative enhancement of local pattern contrast, higher means sharper representations.
(b) \textbf{Channel Effective Rank} is defined as $\mathrm{CER}(x) = \exp(H(p))\cdot C^{-1}$, where $p_i = \lambda_i / \sum_j \lambda_j$ are normalized eigenvalues of the channel covariance and $H(p) = -\sum_i p_i \log p_i$. CER measures the intrinsic dimensionality of channel activations normalized by channel count, indicating diversity of representations, higher means better diversity and less flat response. Both metrics show consistent improvement across all layers with differential kernel modulation, demonstrating enhanced feature sharpness and diversity.}
   \label{fig:3}
\end{figure*}

\noindent\textbf{Lateral Inhibition via Differential Kernel Modulation.}  
Our analysis reveals that self-attention derives its representational power not only from adaptive routing, but also crucially from the inherent \emph{lateral inhibition} dynamics (Eq.~\ref{eq:6}). Thus, we introduce \textit{Differential Kernel Modulation} (DKM), a mechanism that explicitly incorporates lateral inhibition into ATConv. The DKM modulates the final ATConv kernels $\alpha^{\text{ATConv}}$ as follows:

\begin{equation}
\begin{aligned}
\alpha^{\text{ATConv}}_{b,c,u,v} &= \mathbf{K}_{b,c,u,v} - \lambda_c\,\bar{\mathbf{K}}_{b,c}, \\[6pt]
\text{where}\quad 
& \bar{\mathbf{K}}_{b,c} = \tfrac{1}{K^2}\sum_{u,v} \mathbf{K}_{b,c,u,v}, \\[4pt]
\lambda_c &= \sigma(\gamma_c)\in(0,1).
\end{aligned}
\end{equation}

Here, $\gamma \in \mathbb{R}^C$ is a learnable vector that, through the sigmoid function $\sigma(\cdot)$, produces channel-specific inhibition coefficients $\lambda_c \in (0,1)$. $\bar{\mathbf{K}}_{b,c}$ is the spatial mean of the kernel weight. By subtracting $\lambda_c \bar{\mathbf{K}}_{b,c}$ from each kernel position, DKM transforms absolute values into differential signals centered around their spatial average adaptively. This mechanism instantiates the principle of \emph{center-surround antagonism} fundamental to biological vision. In the primary visual cortex, neurons compute responses as central excitation minus weighted surround inhibition, suppressing redundant patterns while preserving salient features. DKM translates this neurobiological principle computationally, with learnable $\lambda_c$ that enables adaptive inhibition profiles for each individual kernel: strong suppression ($\lambda_c \rightarrow 1$) for edge detection, moderate for texture discrimination, and weak ($\lambda_c \rightarrow 0$) for smooth gradient preservation.

A distinguishing characteristic of DKM is its \emph{kernel-wise heterogeneity}. Each kernel maintains independent reference means $\bar{\mathbf{K}}_{b,c}$ and inhibition coefficients $\lambda_c$, establishing diverse competitive dynamics across the feature space. This prevents collapse and promotes functional specialization: high-frequency kernels naturally evolve stronger inhibition to sharpen boundaries, while semantic kernels maintain weaker inhibition for holistic pattern capture. Such differentiation enhances the effective rank of representations by ensuring non-redundant channel responses, while also stabilizing training by preventing degenerate flat activations that can plague deeper layers. Through this biologically-inspired mechanism, DKM transforms standard convolution into a contrast-sensitive operation that balances local detail with global context.

This mechanism is further clarified by the Jacobian structure of DKM with respect to $\mathbf{K}$:
\begin{equation}
\label{eq:lateral_jacobian}
\frac{\partial \alpha^{\text{ATConv}}_{b,c,u,v}}{\partial \mathbf{K}_{b,c,u',v'}}
\;=\;
\delta_{uu'}\delta_{vv'}
\;-\;
\frac{\lambda_c}{K^2}.
\end{equation}

The negative off-diagonal terms $-\lambda_c/K^2$ explicitly encode competition: increasing one weight reduces the influence of others within the same channel. Crucially, the modulation strength is independently determined by each $\lambda_c$, yielding a heterogeneous inhibition landscape that balances sharpening with stability and prevents feature collapse.

\noindent\textbf{Why not Softmax?}  
One might consider applying softmax to enforce competition among kernels. However, it is incompatible with convolution. The simplex probability constraint of softmax removes the essential negative weights for detecting edges and textures, while the exponential scaling drives an extreme winner-take-all effect. It will collapse receptive fields into sparse activations, destroying spatial coherence and leading to training collapse (see Tab.~\ref{tab:9}-(c)).

In contrast, DKM provides the competitive dynamic tailored for Conv. By preserving signed responses, it enables the simultaneous modeling of excitatory and inhibitory patterns crucial for visual contrast. Its differential form ensures smooth gradient flow without saturation, while its adaptive modulation ($\lambda$) automatically discovers heterogeneous suppression strategies across the feature hierarchy: some channels sharpen local discriminative cues, while others capture broader semantic dependencies. Fig.~\ref{fig:3} empirically confirms these effects: DKM enhances representation contrast and increases channel effective rank, jointly yielding sharper and less redundant representations. Together, these properties allow convolution to achieve the competitive dynamics while retaining efficiency.

\noindent\textbf{Complete the ATConv Architecture.} With the kernel $\alpha$ in hand, we now complete the ATConv operator. As established by Eq.~\ref{eq:4}, the adaptive weight alone is not sufficient: if the value space remains fixed, the routing capacity is intrinsically constrained. We therefore leverage a learnable value projection that defines a task-specific basis for ATConv aggregation.
\begin{equation}
\label{eq:value_transform}
\mathbf{V}_{b,c,h,w} \;=\; \sum_{i=1}^{C} \mathbf{W}_{\text{value}}^{(c,i)}\ \mathbf{X}_{b,i,h,w}.
\end{equation}

This projection creates a new, task-optimized feature space (determining what to aggregate) where similar concepts cluster, enabling adaptive routing to perform a far more effective and discriminative aggregation.

In summary, ATConv operates in an efficient depth-wise convolutional framework as follows with the proposed C2K , DKM, and value adaptations.
\begin{equation}
\label{eq:ATConv_routing}
\mathbf{Y}^{\text{ATConv}}_{b,c,h,w}
=
\sum_{u=0}^{K-1}\sum_{v=0}^{K-1}
\alpha^{\text{ATConv}}_{b,c,u,v}(\mathbf{X})\;
\mathbf{V}_{b,c,h+u-p,\;w+v-p},
\end{equation}
where $p=\lfloor K/2\rfloor$. Both the routing weights $\boldsymbol{\alpha}$ and the feature bases $\mathbf{V}$ adapt to input content under principled rules, allowing ATConv to realize adaptive routing with high efficiency. In contrast to Conv’s static kernels and self-attention’s costly quadratic maps, ATConv demonstrates that global adaptivity can be achieved within a compact local processor, redefining how convolutional architectures can embody the expressive power once thought to be exclusive to attention. In the following narrative, we use the term ``$\star$'' to denote the depthwise convolutional operation in Eq.~\ref{eq:ATConv_routing} for simplicity.

Based on the above designs, we define ATConv in Alg.~\ref{alg:1}. For visual clarity, its architecture is illustrated in Fig.~\ref{fig:2}-(b) and (c). Note that following self-attention, we also use an additional linear projection ($\mathbf{W}_{\text{out}}$) before final output.

\begin{algorithm}[t]
\small
    \caption{\textbf{Attentive Convolution (ATConv)}}
    \label{alg:1}
    \renewcommand{\algorithmicrequire}{\textbf{Input:}}
    \renewcommand{\algorithmicensure}{\textbf{Output:}}
    \newcommand{\algorithmicparameters}{\textbf{Parameters:}}
    \newcommand{\PARAMETERS}{\item[\algorithmicparameters]}
    \begin{algorithmic}[1]
    \REQUIRE Input $\mathbf{X}\!\in\!\mathbb{R}^{B\times C\times H\times W}$;\ kernel size $K$
    \PARAMETERS $\operatorname{Conv}_{1\times1}$, $\mathbf{W}_{\text{gen}}$,
    $\boldsymbol{\gamma}_c$, GELU function $\phi(\cdot)$ and Sigmoid function $\sigma(\cdot)$, $\mathbf{W}_{\text{value}}$, $\mathbf{W}_{\text{out}}$.
    \ENSURE  $\mathbf{Y}^{\text{out}}\!\in\!\mathbb{R}^{B\times C\times H\times W}$
    \\ \textcolor{blue!65}{\textbf{// Step-1: Context-to-Kernel Translation}}
    \STATE $\mathbf{F} \leftarrow \operatorname{Conv}_{1\times1}(\mathbf{X})$ \textcolor{gray}{\footnotesize\texttt{\# point-wise conv, $\mathbf{F}\in\mathbb{R}^{B\times C\times H\times W}$}}
    \STATE $\mathbf{Z} \leftarrow \text{AdaAvgPool}_{K\times K}(\mathbf{F})$\textcolor{gray}{\footnotesize\texttt{\# $\mathbf{Z}\in\mathbb{R}^{B\times C\times K\times K}$}}
    \STATE $\mathbf{K} \leftarrow \operatorname{Reshape}\big(\mathbf{W}_{\text{gen}}\cdot  \operatorname{Vec}(\phi(\mathbf{Z}))\big)$ \textcolor{gray}{\footnotesize\texttt{\# $\mathbf{K}\in\mathbb{R}^{B\times C\times K\times K}$}}
    \\ \textcolor{blue!65}{\textbf{// Step-2: Lateral inhibition via Diff. Kernel Modulation}}
    \STATE $\bar{\mathbf{K}}_{b,c} \leftarrow \text{mean}(\mathbf{K}_{b,c,:,:})$\textcolor{gray}{\footnotesize\texttt{\# spatial mean per channel}}
    \STATE $\boldsymbol{\alpha}_{b,c,:,:} \leftarrow \mathbf{K}_{b,c,:,:} - \sigma(\gamma_c)\bar{\mathbf{K}}_{b,c}$ \textcolor{gray}{\footnotesize\texttt{\# per channel competition}}
    \\ \textcolor{blue!65}{\textbf{// Step-3: Convolution with adaptive routing}}
    \STATE $\mathbf{V} \leftarrow \mathbf{W}_{\text{value}} \cdot \mathbf{X}$ \textcolor{gray}{\footnotesize\texttt{\# value projection}}
    \STATE $\mathbf{Y} \leftarrow \boldsymbol{\alpha} \star \mathbf{V}$ \textcolor{gray}{\footnotesize\texttt{\# depthwise conv process}}
    \\ \textcolor{blue!65}{\textbf{// Step-4: Output projection}}
    \STATE $\mathbf{Y}^{\text{out}} \leftarrow \mathbf{W}_{\text{out}} \cdot \mathbf{Y}$ \textcolor{gray}{\footnotesize\texttt{\#output projection}}
    \RETURN $\mathbf{Y}^{\text{out}}$
    \end{algorithmic}
\end{algorithm}

\subsection{Complexity Analysis}
ATConv preserves the dynamic expressivity of self-attention while being substantially more efficient. We quantify these gains from two complementary viewpoints: computational complexity and memory footprint.

\noindent\textbf{Computational Complexity.}
For $\mathbf{X}\!\in\!\mathbb{R}^{B\times C\times H\times W}$ with $N{=}HW$, vanilla self-attention has quadratic complexity:
\begin{equation}
\label{eq:sa_complexity}
\mathcal{O}_{\text{SA}}
\;=\;
\underbrace{\mathcal{O}(NC^2)}_{\text{projections}}
\;+\;
\underbrace{\mathcal{O}(N^2C)}_{\text{attention map}},
\end{equation}
Instead, ATConv consists of (i) context-to-kernel translation, (ii) convolutional aggregation, and (iii) linear projections:
\begin{equation}
\label{eq:atconv_complexity}
\begin{aligned}
\mathcal{O}_{\text{ATConv}}
&=\underbrace{\mathcal{O}(NC^2)}_{\text{context}\rightarrow\text{kernel}}
 \;+\;\underbrace{\mathcal{O}(NK^2C)}_{\text{conv}}
 \;+\;\underbrace{\mathcal{O}(NC^2)}_{\text{projections}}\\[2pt]
&=\;\mathcal{O}\!\big(NC(C{+}K^2)\big)
\;\approx\;\mathcal{O}(NC^2),
\end{aligned}
\end{equation}
typically $K^2\!\ll\!C$. Thus, ATConv is \emph{linear in $N$}, avoiding the quadratic bottleneck of self-attention.

\noindent\textbf{Memory Footprint Analysis.}
Beyond arithmetic complexity, memory is the practical bottleneck for modern deep learning models. Self-attention stores an $N{\times}N$ attention map (with softmax buffers), yielding a quadratic term, whereas ATConv replaces it with a compact bank of dynamic kernels:
\begin{equation}
\label{eq:mem_core}
\begin{aligned}
\text{Mem}_{\text{SA}}   &= \mathcal{O}(BNC) \;+\; \mathcal{O}(BN^2),\\
\text{Mem}_{\text{ATConv}} &= \mathcal{O}(BNC) \;+\; \mathcal{O}(BCK^2).
\end{aligned}
\end{equation}

Furthermore, the routing branch of ATConv is consumed on the fly, so no full $BNC$ activation is persisted; only one large activation (e.g., $V$ or $X$) plus a compact $BCK^{2}$ kernel buffer is retained. By contrast, in addition to the huge $N\times N$ attention maps, SA needs extra $3BNC$ buffers for the three activations Q, K, and V that must be cached for backpropagation. We give a practical example to illustrate the actual memory footprint differences. Generally, let $B{=}32$, $C{=}384$, $H{=}W{=}28$ ($N{=}784$), $K{=}3$, FP16 ($2$ bytes/elt). Using the dominant terms in \eqref{eq:mem_core} with cached $Q/K/V$ will have the following cost:
\begin{equation}
\small
\begin{aligned}
\text{Mem}_{\text{SA}}
&\approx
2\Big(\underbrace{3BNC}_{Q,K,V}\;+\;\underbrace{BN^2}_{\text{attn map}}\Big)
\;\approx\;
92.6~\text{MiB},
\\[4pt]
\text{Mem}_{\text{ATConv}}
&\approx
2\Big(\underbrace{BNC}_{\text{acts}}\;+\;\underbrace{BCK^2}_{\text{dyn kernels}}\Big)
\;\approx\;
18.6~\text{MiB}.
\end{aligned}
\end{equation}

Under this accounting, ATConv reduces per-layer memory by $79.9\%$ at $28{\times}28$ and by $95\%$ at $56{\times}56$, reflecting the removal of the quadratic $BN^{2}$ term. In short, SA’s $\mathcal{O}(BN^{2})$ buffer is replaced by ATConv’s channel-local $\mathcal{O}(BCK^{2})$ buffer, yielding considerable savings while preserving attention-like global dynamics. Even with FlashAttention, which streams computation and avoids storing the explicit $N\times N$ map, narrowing peak memory to $\mathcal{O}(BNC)$, its token2token interactions and memory traffic remain quadratic across tiles. In contrast, ATConv stays linear in $N$ and preserves cache-friendly locality with a $\mathcal{O}(BCK^{2})$ working set, yielding higher throughput and lower memory consumption.

\subsection{Implementation Details}

\noindent\textbf{Overall Architecture of AttNet.} Based on ATConv, we construct AttNet as a family of general-purpose visual backbones, as shown in Fig.~\ref{fig:2}. Specifically, we adopt the ATConv as the spatial operator (token-mixer) and follow the modern ViT architecture to build the model. Following recent best practices \cite{shi2024transnext, hua2022transformer, touvron2023llama, anil2023palm}, we employ the Gated Linear Units (GLU) as a lightweight alternative to the traditional Feed-Forward Network (FFN) for channel mixing. All other components (e.g., double skip connections, Layer Normalization, GELU activation) remain consistent with ViT styles.  In this manner, we develop four variants of AttNet with different budgets, denoted AttNet-T1, -T2, -T3, and -T4. The model sizes and configurations are listed in Tab.~\ref{tab:1}.

\begin{table}[htbp]
  \centering
  \caption{Configuration of four AttNet variants.The number of Blocks and Channels are configured for four stages.}
  \setlength{\tabcolsep}{3.5pt}
  \resizebox{\linewidth}{!}{
    \begin{tabular}{lcccc}
    \cmidrule[0.9pt]{1-5}
     \textbf{Model} & \textbf{\#Params} & \textbf{FLOPs} & \textbf{\#Blocks} & \textbf{\#Channels} \\
    \cmidrule[0.5pt]{1-5}
    AttNet-T1  & 13.7M  & 2.4G   & \makecell[c]{(2, 3, 12, 3)} & \makecell[c]{(48, 96, 224, 384)} \\
    AttNet-T2  & 27.0M  & 5.1G   & \makecell[c]{(3, 3, 16, 3)} & \makecell[c]{(64, 128, 288, 512)} \\
    AttNet-T3  & 49.1M  & 9.4G   & \makecell[c]{(4, 4, 26, 4)} & \makecell[c]{(72, 144, 320, 576)} \\
    AttNet-T4  & 87.3M  & 16.7G  & \makecell[c]{(5, 5, 28, 5)} & \makecell[c]{(96, 192, 384, 768)} \\
    \cmidrule[0.9pt]{1-5}
    \end{tabular}%
    }
  \label{tab:1}%
  \vspace{-2.0em}
\end{table}

\begin{table*}[htbp]
  \centering
  \caption{Comparison with baselines on ImageNet-1K dataset. We replace the attention mechanism in PVT \cite{wang2021pyramid} and Swin \cite{liu2021swin} with ATConv in a drop-in manner to show the variation in accuracy and speed. The Throughput (Thp.) metric is measured on one MI-250X GPU. }
  \resizebox{\linewidth}{!}{
    \begin{tabular}{lccllclccll}
    \cmidrule[0.7pt]{1-5} \cmidrule[0.7pt]{7-11}
    \textbf{Method} & \textbf{\makecell[c]{\#Params\\(M)}} & \textbf{\makecell[c]{FLOPs\\(G)}} & \textbf{\makecell[l]{Thp.\\(fps) \color{BrickRed}{$\uparrow$}}} & \textbf{\makecell[l]{Top-1\\(\%) \color{BrickRed}{$\uparrow$}}} & & \textbf{Method} & \textbf{\makecell[c]{\#Params\\(M)}} & \textbf{\makecell[c]{FLOPs\\(G)}} & \textbf{\makecell[l]{Thp.\\(fps) \color{BrickRed}{$\uparrow$}}} & \textbf{\makecell[l]{Top-1\\(\%) \color{BrickRed}{$\uparrow$}}}\\
    \cmidrule[0.5pt]{1-5} \cmidrule[0.5pt]{7-11}
    PVT-T \cite{wang2021pyramid}& 13.2  & 1.9   & 1701 & 75.1 & & Swin-T \cite{liu2021swin}& 28.3  & 4.5   & 958  & 81.3 \\
   \cellcolor{gray!22}\textbf{ATConv-PvT-T} & \cellcolor{gray!22}\textbf{10.2}  & \cellcolor{gray!22}\textbf{1.8}   & \cellcolor{gray!22}\textbf{2476\color{BrickRed}{\smaller(1.5$\times$)}}  & \cellcolor{gray!22}\textbf{77.5 \color{BrickRed}{\smaller(+2.4)}} & & \cellcolor{gray!22}\textbf{ATConv-Swin-T} & \cellcolor{gray!22}\textbf{28.0}  & \cellcolor{gray!22}\textbf{4.2}   & \cellcolor{gray!22}\textbf{1748\color{BrickRed}{\smaller(1.8$\times$)}}   & \cellcolor{gray!22}\textbf{82.2 \color{BrickRed}{\smaller(+0.9)}}\\
    PVT-S \cite{wang2021pyramid}& 24.5  & 3.8   & 939  & 79.8& & Swin-S \cite{liu2021swin}& 49.6  & 8.8   & 539
  & 83.0  \\
    \cellcolor{gray!22}\textbf{ATConv-PvT-S} & \cellcolor{gray!22}\textbf{18.3}  & \cellcolor{gray!22}\textbf{3.5}   & \cellcolor{gray!22}\textbf{1479\color{BrickRed}{\smaller(1.6$\times$)}} & \cellcolor{gray!22}\textbf{81.7 \color{BrickRed}{\smaller(+1.9)}} & &\cellcolor{gray!22}\textbf{ATConv-Swin-S} & \cellcolor{gray!22}\textbf{47.6}  & \cellcolor{gray!22}\textbf{8.1}   & \cellcolor{gray!22}\textbf{967 \color{BrickRed}{\smaller(1.8$\times$)}} & \cellcolor{gray!22}\textbf{83.6 \color{BrickRed}{\smaller(+0.6)}} \\
    PVT-M \cite{wang2021pyramid}& 44.2  & 6.7   & 590  & 81.2& &Swin-B \cite{liu2021swin}& 87.8  & 15.5  & 364  & 83.5  \\
    \cellcolor{gray!22}\textbf{ATConv-PvT-M} & \cellcolor{gray!22}\textbf{31.9}  & \cellcolor{gray!22}\textbf{6.1}   & \cellcolor{gray!22}\textbf{879 \color{BrickRed}{\smaller(1.5$\times$)}}  & \cellcolor{gray!22}\textbf{82.4 \color{BrickRed}{\smaller(+1.2)}}&   &\cellcolor{gray!22}\textbf{ATConv-Swin-B}  & \cellcolor{gray!22}\textbf{84.2}  & \cellcolor{gray!22}\textbf{14.3}  & \cellcolor{gray!22}\textbf{789 \color{BrickRed}{\smaller(2.2$\times$)}} & \cellcolor{gray!22}\textbf{84.3 \color{BrickRed}{\smaller(+0.8)}}  \\
    PVT-L \cite{wang2021pyramid}& 61.8  & 9.8   & 421  & 81.7&  &Swin-B-384 \cite{liu2021swin} & 87.9  & 47.2  & 105  & 84.5  \\
    \cellcolor{gray!22}\textbf{ATConv-PvT-L} & \cellcolor{gray!22}\textbf{43.3}  & \cellcolor{gray!22}\textbf{9.2}   & \cellcolor{gray!22}\textbf{662    \color{BrickRed}{\smaller(1.6$\times$)}} & \cellcolor{gray!22}\textbf{83.0 \color{BrickRed}{\smaller(+1.3)}} &  &\cellcolor{gray!22}\textbf{ATConv-Swin-B-384} & \cellcolor{gray!22}\textbf{84.3}  & \cellcolor{gray!22}\textbf{42.1}  & \cellcolor{gray!22}\textbf{329 \color{BrickRed}{\smaller(3.1$\times$)}} & \cellcolor{gray!22}\textbf{85.0 \color{BrickRed}{\smaller(+0.5)}} \\
    \cmidrule[0.7pt]{1-5} \cmidrule[0.7pt]{7-11}
    \end{tabular}%
    }
  \label{tab:2}%
  \vspace{-1.5em}
\end{table*}

\begin{figure*}[htpb]
  \centering
   \includegraphics[width=1.0\linewidth]{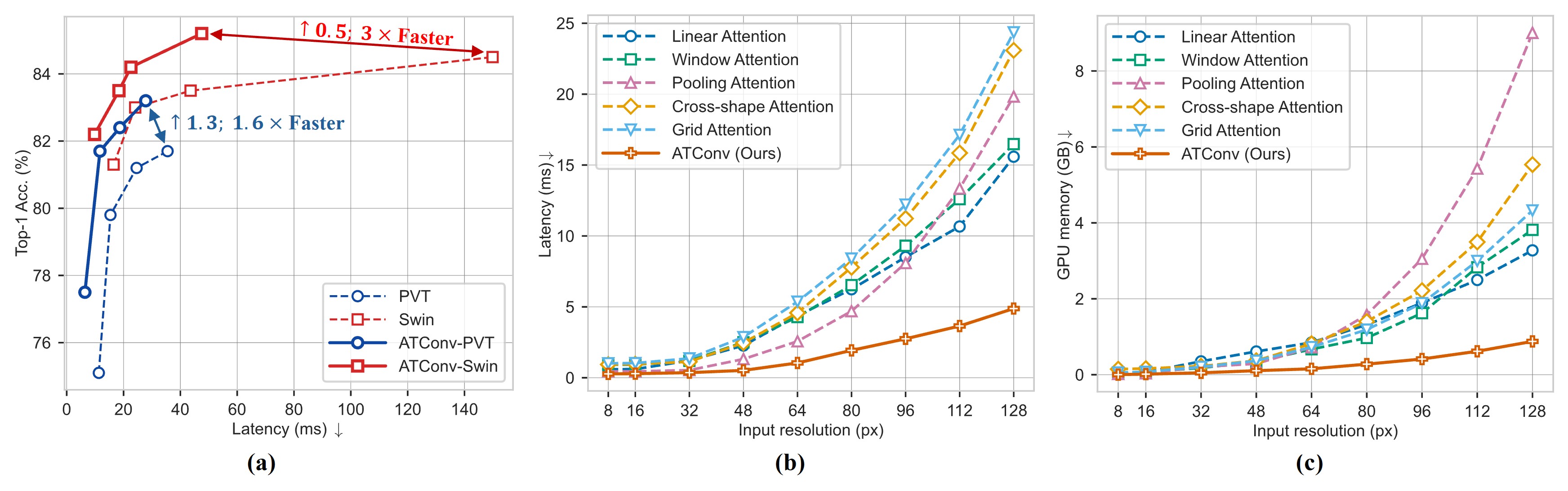}
   \captionsetup{aboveskip=-2.0pt, belowskip=5pt}
   \caption{(a): Latency vs. accuracy curve on PVT \cite{wang2021pyramid} and Swin transformers \cite{liu2021swin}, with their ATConv versions which directly replace the self-attention blocks with ATConv; (b) and (c): Latency and GPU memory consumption in terms of different input resolutions on different operators at the atomic level.}
   \label{fig:4}
   \vspace{-1.5em}
\end{figure*}

\noindent\textbf{ATConv Configuration.} As discussed in Sec.~\ref{sec:2b}, our key insight is to inject adaptive routing and lateral inhibition to achieve higher expressivity,  rather than simply enlarging the receptive field. Although using larger kernels may yield further improvements, we follow our findings and \textit{configure ATConv with a compact $3{\times}3$ kernel size for all AttNet variants}. 

\section{Experiments}
In this section, we first show that ATConv can serve as a drop-in replacement for strong SA mechanisms in image classification, delivering a superior accuracy–efficiency trade-off (Sec.~\ref{sec:A}). We then compare ATConv with state-of-the-art SA variants in computational efficiency (Sec.~\ref{sec:B}). Next, we evaluate AttNet across core vision tasks, including image classification (Sec.~\ref{sec:C}), object detection (Sec.~\ref{sec:D}), and semantic segmentation (Sec.~\ref{sec:E}). We additionally assess robustness of ATConv on cross-modal retrieval (Sec.~\ref{sec:F}) and examine the utility of ATConv in diffusion-based image generation (Sec.~\ref{sec:G}). Finally, we present ablations and analyze architectural design choices (Sec.~\ref{sec:H}).

\begin{table}[t]
  \centering
  \caption{Image classification results on the ImageNet-1K dataset. Top-1 indicates the top-1 accuracy. FLOPs metrics are measured with the input resolution of 224$\times$224.}
  \resizebox{\linewidth}{!}{
    \begin{tabular}{lcccc}
    \cmidrule[1.0pt]{1-5}
    \textbf{Method} & \textbf{Type} & \textbf{\makecell[c]{\#Params\\(M)}} & \textbf{\makecell[c]{FLOPs\\(G)}} & \textbf{\makecell[c]{Top-1\\(\%)}} \\
    \cmidrule[0.5pt]{1-5}
    PVTv2-B1 \cite{wang2022pvt} & ViT   & 13.1  & 2.1   & 78.7 \\
    BiFormer-T \cite{zhu2023biformer} & ViT   & 13.1  & 2.2   & 81.4  \\
    EfficientFormerV2-S2 \cite{li2023rethinking}& ViT   & 12.7  & 1.3   & 82.0  \\
    TransNeXt-Micro \cite{shi2024transnext} & ViT   & 12.8  & 2.7   & 82.5  \\
   \rowcolor{gray!22} \textbf{AttNet-T1} & CNN   & 13.7  & 2.4   & \textbf{82.8}  \\
    \cmidrule[0.5pt]{1-5}
    Swin-T \cite{liu2021swin} & ViT   & 28.3  & 4.5   & 81.3  \\
    PVTv2-B2 \cite{wang2022pvt}& ViT   & 25.4  & 4.0   & 82.1  \\
    Focal-T \cite{yang2021focal} & ViT   & 29.1  & 4.9   & 82.2  \\
    EfficientFormerV2-L \cite{li2023rethinking}& ViT   & 26.1  & 5.2   & 83.5  \\
    STViT-S \cite{huang2022vision} & ViT   & 25.0  & 4.4   & 83.6  \\
    MaxViT-Tiny \cite{tu2022maxvit} & ViT   & 31.0  & 5.6   & 83.6  \\
    BiFormer-S \cite{zhu2023biformer} & ViT   & 25.5  & 4.5   & 83.8  \\
    InLine-CSwin-S \cite{han2024bridging} & ViT   & 33.0  & 4.3   & 83.8  \\
    TransNeXt-Tiny \cite{shi2024transnext}& ViT   & 28.2  & 5.7   & 84.0  \\
    \rowcolor{gray!22} \textbf{AttNet-T2} & CNN   & 27.0  & 5.1   & \textbf{84.4}  \\
    \cmidrule[0.5pt]{1-5}
    Swin-S \cite{liu2021swin}& ViT   & 49.6  & 8.7   & 83.0  \\
    PVTv2-B3 \cite{wang2022pvt}& ViT   & 45.2  & 6.9   & 83.2  \\
    Focal-S \cite{yang2021focal} & ViT   & 51.1  & 9.1   & 83.5  \\
    PVTv2-B4 \cite{wang2022pvt}& ViT   & 62.6  & 10.1  & 83.6  \\
    BiFormer-B \cite{zhu2023biformer}& ViT   & 56.8  & 9.8   & 84.3  \\
    MaxViT-S \cite{tu2022maxvit} & ViT   & 68.9  & 11.7  & 84.5  \\
    TransNeXt-Small \cite{shi2024transnext}& ViT   & 49.7  & 10.3  & 84.7  \\
    STViT-B \cite{huang2022vision} & ViT   & 52.0  & 9.9   & 84.8  \\
     \rowcolor{gray!22}\textbf{AttNet-T3} & CNN   & 49.1  & 9.4   & \textbf{85.3}  \\
    \cmidrule[0.5pt]{1-5}
    Swin-B \cite{liu2021swin}& ViT   & 87.8  & 15.4  & 83.5  \\
    PVTv2-B5 \cite{wang2022pvt}& ViT   & 82.0  & 11.8  & 83.8  \\
    Focal-B \cite{yang2021focal} & ViT   & 89.8  & 16.0  & 83.8  \\
    InLine-CSwin-B \cite{han2024bridging} & ViT   & 73.0  & 14.9  & 84.5  \\
    TransNeXt-Base \cite{shi2024transnext}& ViT   & 89.7  & 18.4  & 84.8  \\
    MaxViT-B \cite{tu2022maxvit} & ViT   & 120.0  & 23.4  & 84.9  \\
    STViT-L \cite{huang2022vision} & ViT   & 95.0  & 15.6  & 85.3  \\
     \rowcolor{gray!22}\textbf{AttNet-T4} & CNN   & 87.3  & 16.7  & \textbf{85.6}  \\
    \cmidrule[1.0pt]{1-5}
    \end{tabular}%
    }
  \label{tab:3}%
  \vspace{-1.5em}
\end{table}%

\begin{table}[t]
  \centering
  \caption{Results on COCO val2017 datasets by drop-in replacing the PVT's pooling attention with ATConv. FLOPs and Throughput (Thp.) are measured under 1280$\times$800 resolution.}
  \setlength{\tabcolsep}{1.8pt}
  \resizebox{\linewidth}{!}{
    \begin{tabular}{l|cc|cccccc}
    \cmidrule[1.0pt]{1-9}
    \multicolumn{9}{c}{\textbf{Mask R-CNN Object Detection on COCO (1$\times$)}} \\
    \cmidrule[0.7pt]{1-9}
    \textbf{Backbone} & \textbf{\makecell[c]{FLOPs\\(G)}} & \textbf{\makecell[c]{Thp.\\(fps)}} & $AP^b$ & $AP^b_{50}$ & $AP^b_{75}$ & $AP^m$ & $AP^m_{50}$ & $AP^m_{75}$ \\
    \cmidrule[1.0pt]{1-9}
    PVT-T \cite{wang2022pvt} & 240   & 54 & 36.7  & 59.2  & 39.3  & 35.1  & 56.7  & 37.3 \\
    InLine-PVT-T \cite{han2024bridging} & 211   & 79 & 40.2  & 62.7  & 43.8  & 37.7  & 59.7  & 40.4 \\
    \rowcolor{gray!22}\textbf{ATConv-PVT-T} & 173   & \textbf{104} & \textbf{42.3}  & \textbf{64.2}  & \textbf{45.6}  & \textbf{39.3}  & \textbf{62.2}  & \textbf{42.4} \\
    \cmidrule[0.7pt]{1-9}
    PVT-S \cite{wang2022pvt} & 305   & 29 & 40.4  & 62.9  & 43.8  & 37.7  & 59.7  & 40.4 \\
    InLine-PVT-S \cite{han2024bridging} & 250   & 47 & 43.4  & 66.4  & 47.1  & 40.1  & 63.1  & 43.3 \\
    \textbf{ATConv-PVT-S} & 215   & \textbf{70} & \textbf{45.9}  & \textbf{68.5}  & \textbf{49.6}  & \textbf{43.3}  & \textbf{65.7}  & \textbf{45.9} \\
    \cmidrule[0.7pt]{1-9}
    PVT-M \cite{wang2022pvt} & 392   & 18 & 42.0    & 64.4  & 45.6  & 39.0    & 61.6  & 42.1 \\
    InLine-PVT-M \cite{han2024bridging} & 301   & 28 & 44.0    & 66.4  & 48.0    & 40.3  & 63.4  & 43.5 \\
    \rowcolor{gray!22}\textbf{ATConv-PVT-M} & 252   & \textbf{45} & \textbf{46.2}  & \textbf{68.8}  & \textbf{50.3}  & \textbf{42.7}  & \textbf{65.8}  & \textbf{45.9} \\
    \cmidrule[0.7pt]{1-9}
    PVT-L \cite{wang2022pvt }& 494   & 12 & 42.9  & 65.0    & 46.6  & 39.5  & 61.9  & 42.5 \\
    InLine-PVT-L \cite{han2024bridging} & 377   & 17 & 45.4  & 67.6  & 49.7  & 41.4  & 64.7  & 44.6 \\
    \rowcolor{gray!22}\textbf{ATConv-PVT-L} & 332   & \textbf{32} & \textbf{47.3}  & \textbf{69.8}  & \textbf{51.8}  & \textbf{43.3}  & \textbf{66.5}  & \textbf{46.4} \\
    \cmidrule[1.0pt]{1-9}
    \end{tabular}%
    }
  \label{tab:4}%
  \vspace{-1.5em}
\end{table}%

\begin{table*}[t]
  \centering
  \caption{Object detection and instance segmentation with Mask R-CNN on COCO val2017 dataset. The FLOPs are measured at resolution 800$\times$ 1280. All models are pretrained on ImageNet-1K. MS schedule means results with multi-scale training.}
  \resizebox{\linewidth}{!}{
    \begin{tabular}{l|cc|cccccc|cccccc}
    \cmidrule[1.0pt]{1-15}
    \multicolumn{1}{c}{\multirow{2}[2]{*}{\textbf{Backbone}}} & \multicolumn{1}{c}{\multirow{2}[2]{*}{\textbf{\makecell[c]{FLOPs\\(G)}}}} & \multirow{2}[2]{*}{\textbf{Type}} & \multicolumn{6}{c|}{\textbf{Mask R-CNN 1$\times$}}             & \multicolumn{6}{c}{\textbf{Mask R-CNN 3$\times$ + MS schedule}} \\
          &       &       & \multicolumn{1}{l}{$AP^b$} & \multicolumn{1}{l}{$AP^b_{50}$} & \multicolumn{1}{l}{$AP^b_{75}$} & \multicolumn{1}{l}{$AP^m$} & \multicolumn{1}{l}{$AP^m_{50}$} & \multicolumn{1}{l}{$AP^m_{75}$} & \multicolumn{1}{l}{$AP^b$} & \multicolumn{1}{l}{$AP^b_{50}$} & \multicolumn{1}{l}{$AP^b_{75}$} & \multicolumn{1}{l}{$AP^m$} & \multicolumn{1}{l}{$AP^m_{50}$} & \multicolumn{1}{l}{$AP^m_{75}$} \\
    \cmidrule[0.7pt]{1-15}
    Swin-T \cite{liu2021swin} & 264  & ViT & 42.2 & 64.4 & 46.2 &  39.1 & 61.6 & 42.0 & 46.0 & 68.2 & 50.2 & 41.6 & 65.1 & 44.8\\
    Focal-T \cite{yang2021focal}& 291   & ViT   & 44.8  & 67.7  & 49.2  & 41.0    & 64.7  & 44.2  & 47.2  & 69.4  & 51.9  & 42.7  & 66.5  & 45.9 \\
    CMT-S \cite{guo2022cmt} & 249 & ViT & 44.6 & 66.8 & 48.9 & 40.7 & 63.9 & 43.4 & 48.3 & 70.4 & 52.3 & 43.7 & 67.7 & 47.1 \\
    UniFormer-S \cite{li2023uniformer} & 269   & ViT   & 45.6  & 68.1  & 49.7  & 41.6  & 64.8  & 45    & 48.2  & 70.4  & 52.5  & 43.4  & 67.1  & 47.0 \\
    \rowcolor{gray!22}\textbf{AttNet-T2} & 225   & CNN   & \textbf{47.3}    & \textbf{69.2}  & \textbf{51.8}  & \textbf{42.6}  & \textbf{66.3}  & \textbf{45.9}  & \textbf{48.6}  & \textbf{70.6}  & \textbf{54.0}  & \textbf{44.2}  & \textbf{67.5}  & \textbf{47.4} \\
    \cmidrule[0.7pt]{1-15}
    Swin-S \cite{liu2021swin} & 354 & ViT &  44.8 & 66.6 & 48.9 & 40.9 & 63.4 & 44.2 & 48.5 & 70.2 & 53.5 & 43.3 & 67.3 & 46.6 \\
    Focal-S \cite{yang2021focal} & 401   & ViT   & 47.4  & 69.8  & 51.9  & 42.8  & 66.6  & 46.1  & 48.8  & 70.5  & 53.6  & 43.8  & 67.7  & 47.2 \\
    DAT-S \cite{xia2022vision} & 378   & ViT   & 47.1  & 69.9  & 51.5  & 42.5  & 66.7  & 45.4  & 49    & 70.9  & 53.8  & 44.0    & 68.0    & 47.5 \\
    UniFormer-B \cite{li2023uniformer} & 399   & ViT   & 47.4  & 69.7  & 52.1  & 43.1  & 66.0    & 46.5  & 50.3  & 72.7  & 55.3  & 44.8  & 69.0    & 48.3 \\
    \rowcolor{gray!22}\textbf{AttNet-T3} & 329   & CNN   & \textbf{49.6}  & \textbf{71.3}  & \textbf{54.2}    & \textbf{44.5}  & \textbf{68.6}  & \textbf{48.3}  & \textbf{50.7}  & \textbf{72.2}  & \textbf{55.5}  & \textbf{45.4}  & \textbf{69.4}  & \textbf{49.2} \\
    \cmidrule[0.7pt]{1-15}
    Swin-B \cite{liu2021swin} & 496   & ViT   & 46.9  & $-$ & $-$ & 42.3  & $-$ & $-$ & 48.5  & 69.8  & 53.2  & 43.4  & 66.8  & 46.9 \\
    Cswin-B \cite{dong2022cswin} & 526   & ViT   & 48.7  & 70.4  & 53.9  & 43.9  & 67.8  & 47.3  & 50.8  & 72.1  & 55.8  & 44.9  & 69.1  & 48.3 \\
    \rowcolor{gray!22}\textbf{AttNet-T4} & 478   & CNN   & \textbf{51.3}  & \textbf{72.6}  & \textbf{56.5}    & \textbf{45.6}  & \textbf{69.8}  & \textbf{49.6}  & \textbf{51.6}  & \textbf{72.9}  & \textbf{56.8}  & \textbf{45.7}  & \textbf{70.2}  & \textbf{49.9} \\
    \cmidrule[1.0pt]{1-15}
    \end{tabular}%
    }
  \label{tab:5}%
\end{table*}%

\begin{table}[htbp]
  \centering
  \caption{Results of semantic segmentation by drop-in replacing the attention in PVT series with ATConv. FLOPs are measured with an input spatial size of 512$\times$2048. Throughput (Thp.) metrics are measured on one MI-250X GPU.}
  \setlength{\tabcolsep}{6.0pt}
  \resizebox{\linewidth}{!}{
    \begin{tabular}{l|ccc|cc}
    \cmidrule[1.0pt]{1-6}
    \multicolumn{6}{c}{\textbf{Semantic Segmentation on ADE20K}} \\
    \cmidrule[0.7pt]{1-6}
    \textbf{Backbone} & \textbf{\makecell[c]{FLOPs\\(G)}} & \textbf{\makecell[c]{Params\\(M)}} & \textbf{\makecell[c]{Thp.\\(fps)}} & \textbf{\makecell[c]{mIoU\\(\%) $\uparrow$}}  & \textbf{\makecell[c]{mAcc\\(\%) $\uparrow$}} \\
    \cmidrule[0.7pt]{1-6}
    PVT-T \cite{wang2021pyramid} & 158   & 17    & 54 & 36.57  & 46.72  \\
    InLine-PVT-T \cite{han2024bridging} & 127   & 16    & 74 & 39.16  & 50.63  \\
    \rowcolor{gray!22}\textbf{ATConv-PVT-T} & 136   & 14    & \textbf{98} & \textbf{42.43}  & \textbf{53.89}  \\
    \cmidrule[0.7pt]{1-6}
    PVT-S \cite{wang2021pyramid} & 225   & 28    & 20 & 41.95  & 53.02  \\
    InLine-PVT-S \cite{han2024bridging} & 168   & 25    & 39 & 42.93  & 54.58  \\
    \rowcolor{gray!22}\textbf{ATConv-PVT-S} & 173   & 23    & \textbf{60} & \textbf{45.81}  & \textbf{56.60}  \\
    \cmidrule[0.7pt]{1-6}
    PVT-L \cite{wang2021pyramid} & 420   & 65    & 12 & 43.99  & 54.62  \\
    InLine-PVT-L \cite{han2024bridging} & 298   & 55    & 16 & 44.71  & 57.17  \\
    \rowcolor{gray!22}\textbf{ATConv-PVT-L} & 292   & 48    & \textbf{29} & \textbf{48.32}  & \textbf{59.18} \\
    \cmidrule[1.0pt]{1-6}
    \end{tabular}%
    }
  \label{tab:6}%
  \vspace{-1.5em}
\end{table}%

\subsection{ATConv as a Drop-in Replacement for Self-Attention}
\label{sec:A}
To rigorously assess whether ATConv can replace self-attention in modern vision backbones, we perform controlled “drop-in replace" experiments on two canonical ViTs: PVT \cite{wang2021pyramid} and Swin \cite{liu2021swin} Transformer, representing pooling-based and window-based visual self-attention designs, respectively. All experiments are conducted on ImageNet-1K under identical training protocols and architectural settings for both baselines and their ATConv replaced variants. We substitute their self-attention modules with ATConv blocks. Besides, we replace the classification token with global average pooling for final feature aggregation, since ATConv’s dense spatial processing renders a dedicated classification token unnecessarily.

Tab.~\ref{tab:2} reports results across multiple model scales. For the PVT family, ATConv delivers substantial gains: PVT-Tiny with ATConv achieves 2.4\% higher Top-1 accuracy while improving the throughput by 1.5$\times$. The Swin Transformer family shows equally compelling results. On the base scale, ATConv-Swin-B improves Top-1 accuracy by 0.8\% while achieving a speedup over 2$\times$, confirming the effectiveness of ATConv in large scales where self-attention typically excels over traditional operators. Across all Swin variants (Tiny through Base), ATConv consistently surpasses window attention in both accuracy and efficiency, with speedups ranging from $1.8\times$ to $2.2\times$. By increasing the resolution to 384px, ATConv can bring further accelerations by $3.1\times$ with a consistent accuracy gain of 0.5\%. Fig.~\ref{fig:4}-(a) further illustrates the latency–accuracy trade-off, where ATConv consistently reduces inference latency while delivering accuracy gains.

The consistent improvements across diverse model scales indicate that ATConv captures fundamental visual structures more efficiently than representative self-attention variants in modern ViTs. The results establish ATConv as a principled convolutional alternative for visual self-attention that advances both accuracy and computational efficiency.

\subsection{Operator-Level Efficiency Comparison}
\label{sec:B}
To assess operator-level efficiency across varying input sizes, we benchmark ATConv against five leading attention mechanisms representing the state-of-the-art accuracy–efficiency trade-off: (1) InLine Linear Attention \cite{han2024bridging}, (2) Cross-Shaped Attention from Cswin \cite{dong2022cswin}, (3) Improved pooling Attention from PVTv2 \cite{wang2022pvt}, (4) Window Attention from Swin \cite{liu2021swin}, and (5) Grid Attention from MaxViT \cite{tu2022maxvit}. For a controlled comparison, each operator is implemented as a standalone module and evaluated directly on input tensors without down-sampling. We fix the channel dimension at $C=128$ and batch size at $B=64$, while progressively increasing spatial resolution from $8 \times 8$ to $128 \times 128$. Latency (ms) and peak GPU memory (MB) are measured on a single MI-250X GPU, with each data point averaged over 10 runs.

\noindent\textbf{Latency.}  
As shown in Fig.~\ref{fig:4}-(b), ATConv consistently outperforms all baselines across resolutions. At the input resolution of 128px, ATConv is $2.1\times$ faster than the InLine Linear Attention and $4.5\times$ faster than Grid Attention. These results demonstrate that ATConv scales more gracefully with spatial resolution, avoiding the quadratic or fragmented computation patterns that burden attention mechanisms. This operator-level evidence confirms that ATConv offers fundamentally higher computational efficiency, particularly on large-resolution inputs where efficiency bottlenecks are most critical.

\noindent\textbf{Memory footprint.} Fig.~\ref{fig:4}-(c) shows greater advantages about memory consumption: ATConv consumes only $1/3$ to $1/10$ of the peak memory required by the compared attention mechanisms at 128px. This huge reduction stems from its elimination of key–query intermediate tensors and the absence of large attention maps, which dominate memory usage in attention. These statistics underscore ATConv's hardware friendliness, allowing deployment in memory-constrained scenarios such as on edge devices, with improved training stability on large-scale servers. 

\noindent\textbf{Summary.} By jointly reducing latency and memory pressure, ATConv provides a compelling efficiency–accuracy balance. Its favorable scaling properties and hardware adaptability establish it as a principled and practical alternative to visual attention, particularly for applications demanding both high throughput and low resource consumption.

\subsection{Image Classification}
\label{sec:C}
Image classification is a fundamental computer vision task, where the goal is to assign a class label to each input image. Many other tasks (e.g., detection and segmentation) build upon networks pretrained on classification as feature extractors. Here, we evaluate AttNet on the ImageNet-1K \cite{deng2009imagenet} dataset and compare with state-of-the-art ViTs.

\noindent\textbf{Experimental Setup.} For a fair comparison, we follow the widely accepted protocols in DeiT \cite{touvron2021training} to train and evaluate our model on the ImageNet-1K dataset. Briefly, AttNet is trained from scratch on ImageNet-1K for 300 epochs, with a total batch size of 4096 distributed across 64 AMD MI-250X GPUs. We use the AdamW \cite{loshchilov2017decoupled} optimizer with a peak learning rate of 4e-3 and a weight decay of 0.05. A 5-epoch linear warm-up is followed by a cosine decay schedule to 1e-5. All training and testing images are resized to 224$\times$224. We adopt commonly accepted augmentations used in DeiT and many other ViTs \cite{shi2024transnext,vasu2023fastvit,wang2022pvt,wu2022p2t}, including RandAugment, MixUp, CutMix, and random erasing. All settings remain consistent for classification across ablations unless explicitly stated.

\noindent\textbf{Experimental Results.} The quantitative results are summarized in Tab.~\ref{tab:3}. For speed comparison, we provide operator-level measurements in Figs.~\ref{fig:4}-(b) and -(c), focusing on the top-5 fastest attention mechanisms selected in Tab.~\ref{tab:3}. Thus, we omit network-level speed metrics here, as they are often confounded by additional architectural factors (e.g., activation functions, depth, and width). In contrast, the atomic operator-level benchmark provided in Fig.~\ref{fig:4} offers a more faithful assessment of the intrinsic efficiency.

As shown in Tab.~\ref{tab:3}, AttNet clearly outperforms state-of-the-art ViTs such as TransNeXt \cite{shi2024transnext} and STViT \cite{huang2022vision}. For instance, AttNet-T1/T2/T3 achieve Top-1 accuracy improvements of 1.4\%, 0.6\%, and 1.0\% over BiFormer, respectively. Compared with recent ViTs that combine sophisticated attention mechanisms with DWConvs, AttNet demonstrates notable advantages while entirely removing the reliance on attention. In particular, AttNet-T2/T3/T4 exceed TransNeXt-Tiny/Small/Base by 0.4\%, 0.6\%, and 0.8\% Top-1 accuracy, respectively, with fewer parameters and FLOPs.

Building upon ATConv, AttNet operates with a purely convolutional architecture to surpass state-of-the-art ViTs. These results demonstrate that by embedding the core advantages of attention into convolutional design, convolutional operators can not only match but also exceed the performance of attention-based models while offering superior efficiency.

\subsection{Object Detection and Instance Segmentation}
\label{sec:D}
Object detection and instance segmentation have long been fundamental and challenging tasks in computer vision. These tasks aim to detect and recognize instances of semantic objects within natural images. In this section, we evaluate AttNet on the MS-COCO \cite{lin2014microsoft} dataset.

\noindent\textbf{Experimental Setup.} Following common practices \cite{liu2021swin, yang2021focal, li2023uniformer, han2024bridging} in the community, we utilize pretrained models on ImageNet-1K as the backbone, integrating Mask R-CNN \cite{he2017mask} and Cascaded R-CNN \cite{cai2019cascade} as the detection and segmentation heads. The models are fine-tuned on the MS-COCO dataset using the AdamW optimizer, following two common experimental configurations: ``1$\times$" (12 training epochs) and ``3$\times$+MS" (36 training epochs with multi-scale training). For comparative analysis, we use the configurations from Swin Transformer and Cswin. All training and evaluations are conducted with MMDetection on a distributed setup using 64 AMD MI-250X GPUs.

\noindent\textbf{Drop-in Evaluation.}
Tab.~\ref{tab:4} summarizes the experimental results for the drop-in replacement of pooling attention mechanisms in PVT with our proposed ATConv. Our ATConv-PVT models significantly outperform the baseline PVT models, demonstrating substantial improvements in both object detection and instance segmentation tasks. Specifically, replacing the pooling attention in PVT-T with ATConv leads to a notable increase in average precision ($AP^b$) from 36.7 to 42.3, while throughput (Thp.) improves from 54 fps to 104 fps. Similarly, for larger PVT variants like PVT-L, ATConv-PVT-L achieves an $AP^b$ improvement from 42.9 to 47.3, alongside a throughput boost from 12 fps to 32 fps. These results underscore the efficacy of ATConv in enhancing both performance and efficiency compared to the standard pooling-based attention mechanisms in vision transformers.

We further compare our ATConv-based approach with InLine attention \cite{han2024bridging}, a state-of-the-art linear attention mechanism. ATConv-PVT consistently outperforms its InLine counterparts with between accuracy and efficiency. For instance, ATConv-PVT-T achieves an $AP^b$ of 42.3, surpassing InLine-PVT-T’s $AP^b$ of 40.2, with a significant throughput improvement from 79 fps (InLine-PVT-T) to 104 fps. This trend holds across other variants: ATConv-PVT-S outperforms InLine-PVT-S with an $AP^b$ of 45.9 versus 43.4, and ATConv-PVT-M similarly exceeds InLine-PVT-M (46.2 vs. 44.0 at $AP^b$).

\noindent\textbf{Comparison with State-of-the-Art Models.}
Tab.~\ref{tab:5} compares ATConv-based backbones with several state-of-the-art models, including Focal Transformer \cite{yang2021focal}, UniFormer \cite{li2023uniformer}, DAT \cite{xia2022vision}, and Cswin \cite{dong2022cswin}. All the compared methods leverage strong attention mechanisms that are powerful for dense prediction. Our AttNet consistently outperforms these models across all metrics with efficient 3$\times$3 spatial kernels. Specifically, AttNet-T2 achieves an $AP^b$ of 47.3, surpassing Focal-T (44.8) and UniFormer-S (45.6). Additionally, AttNet-T4 outperforms Cswin-B, with a 2.6-point higher $AP^b$ and a 1.7-point higher $AP^m$ under the $1\times$ training schedule.

\begin{table*}[t]
  \centering
  \setlength{\abovecaptionskip}{2pt}
  \caption{Comparison on LLCM \cite{LLCM} and VCM-HITSZ \cite{vcm} benchmarks for visible-infrared image retrieve in the cross-modality setting. 
  $r$ denotes rank-$k$ accuracy, mAP denotes mean Average Precision, and mINP denotes mean Inverse Negative Penalty.}
   \setlength{\tabcolsep}{7.5pt}
  \resizebox{\linewidth}{!}{
    \begin{tabular}{l|c|ccccc|ccccc}
    \cmidrule[1.0pt]{1-12}
    \multicolumn{1}{l|}{\multirow{2}[2]{*}{\textbf{Method}}} & 
    \multicolumn{1}{c|}{\multirow{2}[2]{*}{\textbf{\makecell[c]{Params\\(M)}}}} & 
    \multicolumn{5}{c|}{\textbf{LLCM} \cite{LLCM}} & 
    \multicolumn{5}{c}{\textbf{VCM-HITSZ} \cite{vcm}} \\
    & & $r$=1 & $r$=5 & $r$=20 & mAP & mINP & $r$=1 & $r$=5 & $r$=20 & mAP & mINP \\
    \cmidrule[0.7pt]{1-12}
    EfficientFormerV2-S2 \cite{li2023rethinking}& 12.6  & 42.51 & 66.49 & 84.88 & 50.15 & 46.62 & 34.71 & 54.14 & 71.31 & 24.14 & 7.88 \\
    PVTv2-B1 \cite{wang2022pvt}           & 13.1  & 48.36 & 71.43 & 88.02 & 55.71 & 52.37 & 51.35 & 68.61 & 81.26 & 37.45 & 15.24 \\
    \rowcolor{gray!22} \textbf{AttNet-T1}           & 13.7  & \textbf{50.85} & \textbf{73.28} & \textbf{88.67} & \textbf{57.82} & \textbf{54.49} & \textbf{56.50} & \textbf{72.77} & \textbf{83.51} & \textbf{42.47} & \textbf{18.00} \\
    \cmidrule[0.7pt]{1-12}
    ResNet-50  \cite{he2016deep}          & 25.6  & 36.47 & 59.36 & 79.21 & 43.54 & 39.80 & 37.30 & 56.20 & 71.75 & 23.54 & 5.71 \\
    ConvNeXt-Tiny  \cite{liu2022convnet}      & 29.0  & 36.56 & 60.15 & 79.76 & 44.04 & 40.62 &  5.04 & 14.38 & 29.56 &  3.76 & 0.83 \\
    Swin-Tiny \cite{liu2021swin}           & 29.0  & 42.06 & 66.90 & 85.32 & 50.10 & 46.80 & 51.55 & 67.56 & 79.97 & 38.55 & 16.70 \\
    EfficientFormerV2-L \cite{li2023rethinking} & 26.1  & 45.83 & 69.39 & 86.57 & 53.25 & 49.75 & 46.62 & 63.53 & 77.53 & 33.98 & 13.30 \\
    Focal-Tiny \cite{yang2021focal}          & 29.1  & 48.55 & 71.38 & 87.73 & 55.72 & 52.35 & 59.60 & 75.20 & 85.17 & 46.42 & 21.21 \\
    PVTv2-B2 \cite{wang2022pvt}          & 25.4  & 48.82 & 71.56 & 87.80 & 56.03 & 52.70 & 58.60 & 73.73 & 84.25 & 44.07 & 19.99 \\
    MaxViT-Tiny \cite{tu2022maxvit}          & 31.0  & 50.00 & 72.62 & 88.41 & 57.15 & 53.79 & 59.60 & 74.15 & 84.18 & 47.09 & 22.50 \\
    STViT-Small \cite{huang2022vision}         & 25.0  & 50.18 & 72.78 & 88.65 & 57.06 & 53.51 & 54.08 & 70.09 & 82.22 & 41.56 & 18.35 \\
    \rowcolor{gray!22}\textbf{AttNet-T2} & 27.0  & \textbf{52.41} & \textbf{74.85} & \textbf{89.60} & \textbf{59.54} & \textbf{56.32} & \textbf{62.74} & \textbf{76.07} & \textbf{85.75} & \textbf{50.07} & \textbf{25.46} \\
    \cmidrule[1.0pt]{1-12}
    \end{tabular}%
  }
  \label{tab:7}
  \vspace{-1.5em}
\end{table*}

\begin{table*}[htbp]
  \centering
  \setlength{\abovecaptionskip}{2pt}
  \caption{FID Comparisons with vanilla SiT and REPA on FFHQ \cite{karras2019style} at 512$\times$512 resolution and ImageNet-1K \cite{karras2019style} at 256$\times$256 resolution. For FFHQ,  we do not use classifier-free guidance
 (CFG) and sampling using unconditional generation settings. For ImageNet-1K, we sampling use a consistent CFG Scale of 1.8 without additional scheduling. $\downarrow$ denotes lower the better.}
  \setlength{\tabcolsep}{5.5pt}
  \resizebox{\linewidth}{!}{
    \begin{tabular}{r|cc|ccccc|cccccc}
    \cmidrule[1.0pt]{1-14}
    \multirow{2}[2]{*}{Model} & \multirow{2}[2]{*}{\makecell[c]{\#Params\\(M)}} & \multirow{2}[2]{*}{Iter.} & \multicolumn{5}{c|}{\textbf{FFHQ \cite{karras2019style} 512$\times$512}}      & \multicolumn{6}{c}{\textbf{ImageNet-1K \cite{deng2009imagenet} 256$\times$256 with REPA \cite{yu2024representation}}} \\
          &       &       & Lat.$\downarrow$ & FID$\downarrow$  & sFID$\downarrow$ & Pre.$\uparrow$ & Rec.$\uparrow$ & Lat.$\downarrow$ & FID$\downarrow$  &sFID$\downarrow$ & IS$\uparrow$ & Pre.$\uparrow$ & Rec.$\uparrow$ \\
    \cmidrule[0.7pt]{1-14}
    SiT-B/2 \cite{peebles2023scalable} & 130.32  & 400K  & 139.64  & 10.42  & 26.45  & 0.60  & 0.47  & 31.17  & 8.01  & 5.78  & 147.72  & 0.70  & 0.57  \\
    SiT-Hybrid-B/2 & 126.78  & 400K  & 112.43  & \textbf{10.09}  & \textbf{19.76}  &\textbf{ 0.62}  & 0.53  & 27.41  & 7.17  & 5.44  & 153.78  & \textbf{0.72}  & 0.59  \\
    \rowcolor{gray!22} SiT-ATConv-B/2 & 123.24  & 400K  & \textbf{103.14}  & 10.31  & 23.76  & 0.63  & \textbf{0.52}  & \textbf{25.45} & \textbf{7.15}  & \textbf{5.10}  & \textbf{149.23}  & 0.73  & \textbf{0.58}  \\
    \cmidrule[0.7pt]{1-14}
    SiT-L/2 \cite{peebles2023scalable} & 457.84  & 400K  & 412.71  & 9.36  & 24.75  & 0.62  & 0.51  & 97.44  & 2.12  & 4.87  & 265.50  & 0.79  & 0.56  \bigstrut[t]\\
     SiT-Hybrid-L/2 & 445.21  & 400K  & 375.19  & \textbf{7.74}  & 16.83  & 0.65  & 0.57  & 89.71  & 1.97  & 4.66  & 262.28  & \textbf{0.81}  & 0.60  \\
     \rowcolor{gray!22}SiT-ATConv-L/2 & 432.68  & 400K  & \textbf{341.01}  & 7.99  & \textbf{16.13}  & \textbf{0.66}  & \textbf{0.59}  & \textbf{82.22}  & \textbf{1.95}  & \textbf{4.61}  & \textbf{267.75}  & \textbf{0.81}  & \textbf{0.61}  \\
    \cmidrule[0.7pt]{1-14}
    SiT-XL/2 \cite{peebles2023scalable} & 674.83  & 400K  & 647.75  & 8.87  & 18.48  & 0.65  & 0.56  & 139.44  & 1.97  & 4.76  & 282.33  & 0.79  & 0.58  \\
     SiT-Hybrid-XL/2 & 656.26  & 400K  & 565.96  & \textbf{7.72}  & 17.91  & 0.67  & 0.60  & 121.52  & 1.86  & 4.76  & 290.61  & 0.82  & 0.61 \\
      \rowcolor{gray!22}SiT-ATConv-XL/2 & 637.68  & 400K  & \textbf{501.39}  & 7.88  & \textbf{17.12}  & \textbf{0.69} & \textbf{0.61}  & \textbf{113.05}  & \textbf{1.82}  & \textbf{4.71}  & \textbf{291.17}  & \textbf{0.83}  & \textbf{0.62}  \\
    \cmidrule[1.0pt]{1-14}
    \end{tabular}%
    }
  \label{tab:8}%
  \vspace{-1.0em}
\end{table*}%

These results highlight the competitive performance of AttNet against ViTs in dense prediction tasks. In particular, with only a 3$\times$3 kernel, AttNet achieves performance on par with various attention variants with larger receptive fields. This underscores the key contribution of our ATConv, which efficiently encodes global scene understanding into local processing rules with the context-to-kernel translation. It provides an effective and computationally efficient alternative to attention mechanisms for object understanding.

\subsection{Semantic Segmentation}
\label{sec:E}
Semantic segmentation involves assigning a semantic label to each pixel in an image, making it one of the most critical tasks in computer vision that asserts the dense prediction capacity of foundation models. We benchmark the proposed ATConv on the ADE20K dataset for semantic segmentation. 

\noindent\textbf{Setup.}  We evaluate the performance of ATConv when integrated into the PVT series for semantic segmentation on the ADE20K \cite{zhou2017scene} dataset. We replace the attention mechanism in the PVT backbone with the proposed ATConv and measure the performance in terms of mean Intersection over Union (mIoU) and mean Accuracy (mAcc). We employ SemanticFPN as segmentation heads and follow the protocols in \cite{han2024bridging, wang2021pyramid} for fair comparisons.

\noindent\textbf{Experimental results.} As shown in Tab.~\ref{tab:6}, the results indicate that ATConv consistently outperforms both the PVT and InLine variants in terms of segmentation accuracy, while also achieving superior throughput. For example, ATConv-PVT-T improves the mIoU to 42.43\% and mAcc to 53.89\%, surpassing the InLine-PVT-T model, which achieves a mIoU of 39.16\% and mAcc of 50.63\%. Notably, ATConv-PVT-T achieves a remarkable throughput of 98 fps, significantly higher than both PVT-T (54 fps) and InLine-PVT-T (74 fps).

In the larger model configurations, ATConv continues to show strong improvements. For instance, ATConv-PVT-S outperforms InLine-PVT-S with a mIoU of 45.81\% and mAcc of 56.60\%, along with a throughput of 60 fps, compared to 39 fps for InLine-PVT-S. Similarly, ATConv-PVT-L achieves a mIoU of 48.32\% and mAcc of 59.18\%, with an impressive throughput of 29 fps, outperforming InLine-PVT-L, which has a mIoU of 44.71\% and mAcc of 57.17\% at 16 fps.

These results highlight the effectiveness of ATConv in enhancing both the segmentation accuracy and efficiency of PVT series, demonstrating its potential as a drop-in replacement of attention mechanisms in semantic segmentation tasks.

\subsection{Evaluation on Cross-domain Robustness}
\label{sec:F}
We evaluate ATConv's robustness on cross-modality understanding using the challenging LLCM \cite{LLCM} and VCM-HITSZ \cite{vcm} datasets for visible--infrared retrieval. This task requires establishing reliable latent correspondence between visible and infrared modalities under huge imaging spectral discrepancies. While self-attention typically demonstrates superior representational robustness over convolution, we show that ATConv achieves better robustness to various self-attention variants.

\noindent\textbf{Setup.} All methods are trained on LLCM \cite{LLCM} and VCM-HITSZ \cite{vcm} using ImageNet-1K pretrained models, following the standard image retrieval training and evaluation protocol in \cite{AGW}. For fair comparison, all images are resized to $224 \times 224$ to meet the strict input resolution requirements of some attention-based baselines. Following common practices \cite{CAJ,MUN,DNS,L2RW}, we comprehensively report the rank at $r$, mean Average Precision (mAP), and mean Inverse Negative Penalty (mINP) \cite{AGW} as evaluation metrics (all metrics are higher the better).

\begin{figure*}[t]
  \centering
   \includegraphics[width=1.0\linewidth]{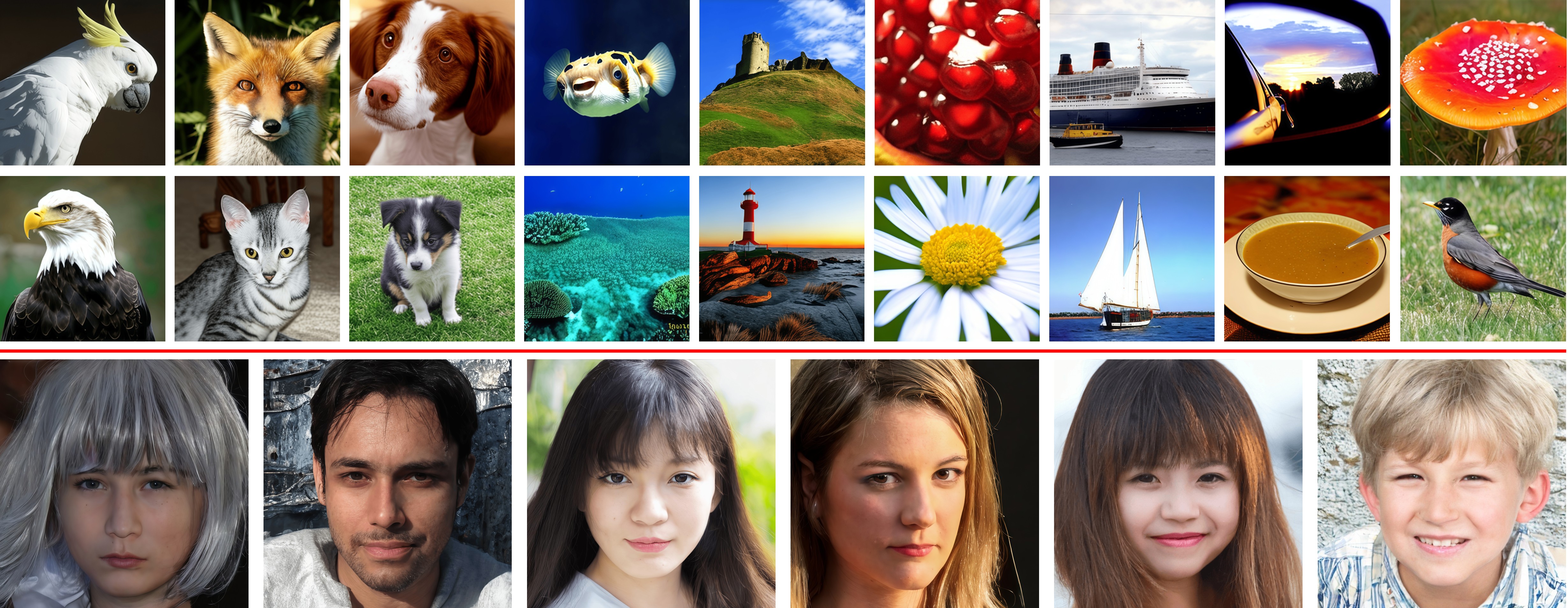}
    \setlength{\abovecaptionskip}{2pt}\caption{Representative images generated by SiT-ATConv-XL/2, where all attention mechanisms are replaced with ATConv to form a pure convolutional generative architecture. The top two rows present natural images synthesized on the ImageNet-1K dataset using classifier-free guidance ($w=4.0$). The bottom row shows facial images unconditionally synthesized on the FFHQ dataset. We show results at 400K steps, more training steps will yield better quality.}
   \label{fig:5}
\end{figure*}

\begin{figure}[t]
  \centering
   \includegraphics[width=0.9\linewidth]{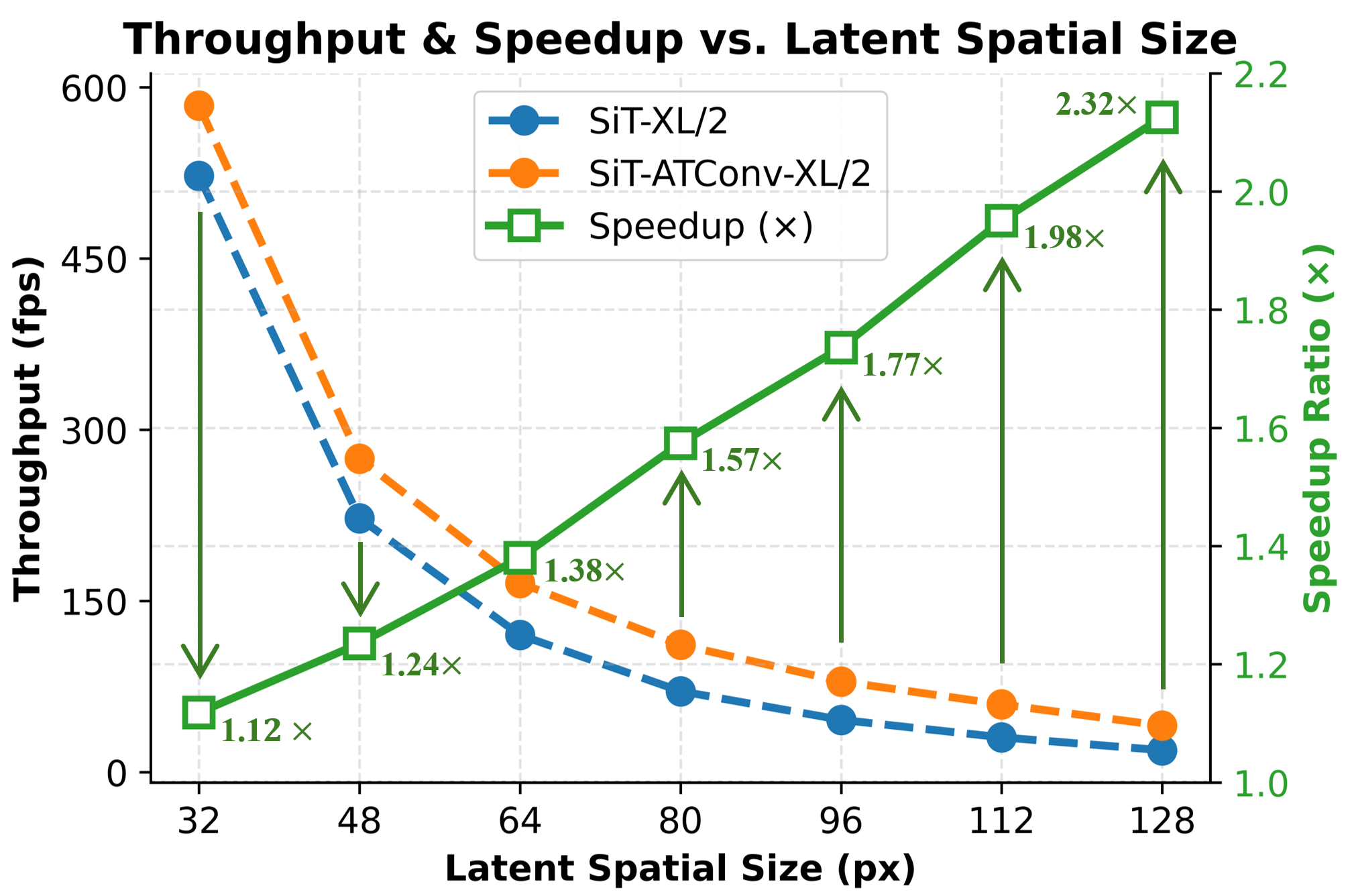}
    \setlength{\abovecaptionskip}{2pt}\caption{Comparison of throughput scaling with different latent sizes. ATConv achieves greater efficiency advantage at larger latent resolutions.}
    \vspace{-1.0em}
   \label{fig:6}
\end{figure}

\noindent\textbf{Results.} Tab.~\ref{tab:7} demonstrates AttNet's superior robustness on cross-modality understanding. AttNet-T1 achieves the mAP of 57.82\%/42.47\% on LLCM/VCM-HITSZ, surpassing PVTv2-B1 by 2.11\%/5.02\% with comparable parameters. AttNet-T2 establishes SOTA mAP with 59.54\%/50.07\% mAP on LLCM/VCM-HITSZ, significantly outperforming the attention-based solutions MaxViT-Tiny (57.15\%/47.09\%) and STViT-Small (57.06\%/41.56\%). AttNet also consistently achieves the best rank-$r$ and mINP metrics for feature-based retrieval, demonstrating superior capacity for learning robust representations in heterogeneous latent spaces.

The robustness of AttNet comes from adaptive routing and lateral inhibition, the former adaptively aggregates semantically relevant features across modalities despite spectral gaps, while the latter injects competitive dynamics to suppress noise and amplify modality-shared cues. These mechanisms enable AttNet to achieve better efficiency-robustness tradeoff, particularly evident in the 8.51\% improvement in mAP over the robust STViT-Small on the VCM-HITSZ benchmark.

\subsection{ATConv for Diffusion Image Generation}
\label{sec:G}
Diffusion models \cite{peebles2023scalable, croitoru2023diffusion, he2025diffusion, yu2024representation} have become the cornerstone of visual AIGC, achieving an unprecedented quality of visual synthesis. Unlike traditional discriminative tasks (e.g., classification and detection) that identify patterns from existing images, diffusion generation demands constructing coherent visual structures from noise, demanding exceptional capacity for both global semantic consistency and fine-grained detail synthesis. This complexity has established transformer \cite{peebles2023scalable, yu2024representation, cao2025difftf, li2024latent} as the dominant architecture.

We challenge this assumption by investigating whether ATConv, as an efficient convolutional operator, can match the generation quality of self-attention. Departing from conventional research limited to discriminative tasks, in this paper, we directly test ATConv in the diffusion-based image generation task where attention has been deemed irreplaceable.

\noindent\textbf{Setup.} We evaluate ATConv in two scenarios: (1) unconditional 512$\times$ 512 facial image generation on FFHQ \cite{karras2019style} using SiT \cite{peebles2023scalable} as the baseline, and (2) conditional 256$\times$256 natural image generation on ImageNet-1K \cite{deng2009imagenet} using SiT \cite{peebles2023scalable} with REPA \cite{yu2024representation} as the baseline. Building upon the baseline, we create SiT-ATConv variants by replacing all attention blocks in SiT with our ATConv. Besides, we further present hybrid variants (SiT-Hybrid) that replace only even-numbered attention blocks with ATConv for an interleaved architecture. All models maintain identical training protocols and evaluation metrics as their baselines \cite{peebles2023scalable, yu2024representation} to ensure fair comparisons.

\noindent\textbf{Results.} Tab.~\ref{tab:8} presents quantitative results, revealing that ATConv exceeds self-attention for diffusion-based image generation with better generation quality and efficiency. On FFHQ at 512$\times$512 resolution, SiT-ATConv-XL/2 achieves better FID (7.88 vs 8.87) with 22.6\% latency reduction compared to the vanilla SiT-XL, while improving both precision (0.69 vs 0.65) and recall (0.61 vs 0.56). Even remarkably, on ImageNet-1K with the REPA training technique, our full ATConv variants consistently outperform attention-based baselines: SiT-ATConv-XL/2 achieves significantly better FID (1.82 vs 1.97), higher IS (291.17 vs 282.33), and superior precision-recall trade-offs, all while reducing latency by 19\%. These improvements scale consistently across all model sizes, with SiT-ATConv-B/2 reducing FID from 8.01 to 7.15 on ImageNet while accelerating inference by 18\%. Besides, we show that by replacing only half of the attention with ATConv, the SiT-Hybrid variants can also produce remarkable improvements.

The consistent superiority of ATConv reveals a fundamental alignment with the denoising objective. ATConv constructs content-adaptive, \textit{signed} spatial kernels with DKM that enforces mean-shifted filtering, naturally suppressing uninformative DC components while enhancing contrastive cues critical for $\epsilon$-prediction. The signed kernels enable simultaneous excitatory and inhibitory responses that preserve fine details during iterative denoising, while DKM's competition dynamics stabilize gradients across varying noise levels. This inductive bias particularly benefits early denoising steps, explaining the observed superior sample quality.

Fig.~\ref{fig:5} visually confirms these quantitative gains, with SiT-ATConv-XL/2 generating sharp facial details and coherent natural images despite using a purely convolutional architecture. These results suggest that the ATConv can serve as a new foundational operator for diffusion-based visual generation with both higher quality and efficiency. Fig.~\ref{fig:6} further shows that the speed advantage of SiT-ATConv-XL to SiT-XL grows with increasing latent resolution. We hope that the excellent performance of ATConv can inspire the development of future generative architectures with higher efficiency and quality.

\subsection{Ablation Studies}
\label{sec:H}
We conduct ablation studies to thoroughly examine the effectiveness of ATConv designs and the influence of different hyperparameter settings. All experiments are performed on the ImageNet-1K dataset following the protocol described in Sec.~\ref{sec:C}, with AttNet-T2 (27M parameters) chosen as the baseline. We report results in the following three aspects.  

\noindent\textbf{Effect of kernel size.}  
Tab.~\ref{tab:9}-(a) investigates the impact of varying kernel sizes in ATConv. When using a uniform kernel size across all stages, the $3\times3$ configuration emerges as the most cost-effective, offering the best balance between accuracy and speed. Larger kernels ($5\times5$ and $7\times7$) yield marginal accuracy gains (0.04\% and 0.11\%) but at a substantial cost in efficiency. Hierarchical configurations further reveal that simply enlarging receptive fields does not guaranty better performance. For instance, setting $[7,7,5,3]$ across the four stages achieves the highest accuracy (84.53\%), slightly outperforming the uniform $7\times7$ setting. This highlights that tailoring kernels to the intrinsic spatial properties of each stage is more effective than blindly enlarging them.  

From another perspective, the accuracy improvements from larger ATConv kernels are modest compared to the drastic gains typically observed when scaling DWConv kernels. This suggests that ATConv, through its efficient integration of adaptive routing and lateral inhibition, already extracts rich visual representations using compact $3\times3$ kernels. In effect, ATConv breaks the traditional paradigm of pursuing ever-larger receptive fields and instead demonstrates a principled, efficient approach to spatial modeling.

\begin{table}[t]
\vspace{-0.5em}
  \centering
  \setlength{\abovecaptionskip}{1pt}
  \caption{Quantitative results of Ablation studies.}
  \setlength{\tabcolsep}{5.5pt}
  \resizebox{\linewidth}{!}{
    \begin{tabular}{l|cccc}
    \cmidrule[1.0pt]{1-5}
    \multicolumn{5}{c}{\textbf{(a) Ablation on Kernel Size of ATConv}} \\
    \textbf{Kernel Size} & \textbf{\makecell[c]{Params\\(M)}} & \textbf{\makecell[c]{FLOPs\\(G)}} & \textbf{\makecell[c]{Thp.\\(fps)}}  & \textbf{\makecell[c]{Top-1\\(\%)}} \\
    \cmidrule[0.7pt]{1-5}
    Unitary $3\times3$ (default) & 27.01 & 5.11  & 1128  & 84.41  \\
    Unitary $5\times5$ & 27.02 & 5.15  & 953   & 84.45  \\
    Unitary $7\times7$ & 27.07 & 5.20  & 831   & 84.52  \\
    Hierarchical $[7, 5, 3, 3]$ & 27.02 & 5.14  & 947
   & 84.46  \\
    Hierarchical $[7, 5, 5, 3]$ & 27.03 & 5.16  & 893
   & 84.51  \\
     \rowcolor{gray!22}Hierarchical $[7, 7, 5, 3]$ & 27.03 & 5.17  & 865   & 84.53  \\
    \cmidrule[0.7pt]{1-5}
    \multicolumn{5}{c}{\textbf{(b) Ablation on Different Token Mixers}} \\
    \textbf{Operator} & \textbf{\makecell[c]{Params\\(M)}} & \textbf{\makecell[c]{FLOPs\\(G)}} & \textbf{\makecell[c]{Thp.\\(fps)}}  & \textbf{\makecell[c]{Top-1\\(\%)}} \\
    \cmidrule[0.7pt]{1-5}
     \rowcolor{gray!22}Default ($3\times3$ ATConv) & 27.01 & 5.11  & 1128  & 84.41  \\
    $\rightarrow 3\times3$ Conv & 40.04 & 7.35  & 787   & 81.81  \\
    $\rightarrow 3\times3$ DWConv & 20.52 & 3.98  & 1539  & 78.06  \\
    $\rightarrow 5\times5$ DWConv & 20.63 & 4.01  & 1311  & 79.11  \\
    $\rightarrow 7\times7$ DWConv & 20.79 & 4.06  & 1168  & 80.31  \\
    \cmidrule[0.7pt]{1-5}
    $\rightarrow$  Hydra Attention \cite{bolya2022hydra} & 29.18 & 5.47  & 772   & 79.83  \\
    $\rightarrow$  InLine Attention \cite{han2024bridging} & 31.63 & 5.63  & 684   & 83.61  \\
    $\rightarrow$  RankAug Attention \cite{fan2025breaking} & 31.54 & 6.11  & 543   & 83.67  \\
    \cmidrule[0.7pt]{1-5}
    \multicolumn{5}{c}{\textbf{(c) Ablation on Building ATConv from DWConv}} \\
    \textbf{Operator Config.} & \textbf{\makecell[c]{Params\\(M)}} & \textbf{\makecell[c]{FLOPs\\(G)}} & \textbf{\makecell[c]{Thp.\\(fps)}}  & \textbf{\makecell[c]{Top-1\\(\%)}} \\
    \cmidrule[0.7pt]{1-5}
    $3\times3$ DWConv & 20.52  & 3.98  & 1539  & 78.06  \\
     \rowcolor{gray!22}$+$ Kernel Generator & 22.64  & 4.37  & 1457  & 80.94  \\
    $+$ Last Linear Proj. & 24.82  & 4.74  & 1338  & 81.65  \\
     \rowcolor{gray!22}$+$ Value Proj. & 27.00  & 5.11  & 1190  & 83.17 \\
    \cmidrule[0.7pt]{1-5}
    $+$ Softmax on $\mathbf{K}$ & 27.00  & 5.11  & 1139  & $-$ \\
    $\rightarrow$ Kernel Diff. on $\mathbf{K}$ & 27.00  & 5.11  & 1168  & 82.80 \\
     \rowcolor{gray!22}$\rightarrow$ Diff. Modulation on $\mathbf{K}$ & 27.01  & 5.11  & 1128  & 84.41 \\
    \cmidrule[1.0pt]{1-5}
    \end{tabular}%
    }
  \label{tab:9}%
  \vspace{-1.0em}
\end{table}%

\noindent\textbf{Replace ATConv with alternative operators.}  
Tab.~\ref{tab:9}-(b) compares ATConv with its alternative operators by replacing ATConv in AttNet-T2 with conventional convolutions or attention mechanisms. Replacing ATConv with vanilla Conv or DWConv (of various kernel sizes) leads to clear drops in Top-1 accuracy, underscoring ATConv’s superior accuracy–efficiency trade-off and its stronger representational capacity.  

We also benchmark three leading attention mechanisms with linear spatial complexity: Hydra Attention, InLine Attention, and RankAug Attention. As shown in Tab.~\ref{tab:9}-(b), all three underperform ATConv in both accuracy and efficiency. For example, InLine and RankAug attention trail ATConv by 0.80\% and 0.74\% in accuracy, while running 1.64$\times$ and 2.0$\times$ slower, respectively. These results confirm that by uniting the adaptivity of attention with the inductive bias of convolution, ATConv surpasses both traditional convolutions and advanced attention-based token mixers.  

\noindent\textbf{Roadmap from DWConv to ATConv.}  
Finally, Tab.~\ref{tab:9}-(c) illustrates the progressive transformation from a standard $3\times3$ DWConv into ATConv, with performance gains measured at each step. Introducing the kernel generator, which converts static kernels into dynamic ones, yields a large accuracy boost (+2.88\%) with negligible overhead. Adding a value projection and a final linear projection, thereby enabling adaptive routing coupled with dynamic kernels—further improves accuracy by 0.71\% and 1.52\%, respectively.  

To validate the effectiveness of differential kernel modulation mechanism towards injecting the lateral inhibition attribute into convolutional calculation, we compare three alternatives: applying softmax on kernels (which causes training collapse), using a classic central difference \cite{su2021pixel} (which degrades performance due to indiscriminate suppression of low-frequency signals), and our differential kernel modulation. As reported in Tab.~\ref{tab:9}-(c), the latter delivers a further accuracy gain of +1.24\% with minimal computational cost. These results demonstrate that each component of ATConv contributes meaningfully and consistently, collectively forging a highly effective visual operator.  

\section{Conclusion}
This paper presents the first systematic identification of adaptive routing and lateral inhibition as the essential principles driving the success of attention mechanisms. Leveraging these insights, we introduce Attentive Convolution (ATConv), a purely convolutional operator that inherits these properties while preserving the efficiency and visual inductive biases of convolution. Our experiments show that ATConv not only surpasses leading attention mechanisms and Conv-attention hybrids in both accuracy and efficiency, but also establishes a scalable foundation for future architectures. We hope that this work will pave a new path for the evolution of convolutional architectures, bridging the gap with attention and inspiring future research in efficient visual modeling. We acknowledge two primary limitations of our current work. First, ATConv has not been explored in autoregressive settings, which constrains its applicability to next-token prediction paradigms prevalent in large language models. Second, while ATConv demonstrates computational advantages over most existing operators, its adaptive design introduces overhead compared to vanilla convolution. Future work will focus on extending ATConv to autoregressive frameworks and optimizing its efficiency through custom CUDA implementations.

\bibliographystyle{IEEEtran}
\bibliography{reference}

\end{document}